\pdfoutput=1
\documentclass{article} %
\usepackage{iclr2027_conference,times}
\usepackage[T1]{fontenc} %

\newif\ifpreprint
\preprinttrue
\ifpreprint\iclrfinalcopy\fi

\usepackage{amsmath,amsfonts,bm}

\def\eqref#1{equation~\ref{#1}}

\def\1{\bm{1}}

\DeclareMathAlphabet{\mathsfit}{\encodingdefault}{\sfdefault}{m}{sl}
\SetMathAlphabet{\mathsfit}{bold}{\encodingdefault}{\sfdefault}{bx}{n}

\usepackage{hyperref}
\usepackage{url}
\usepackage{amsmath,amssymb,amsthm}
\usepackage{booktabs}
\usepackage{graphicx}
\usepackage{xcolor}
\usepackage{caption}
\newtheorem{proposition}{Proposition}
\newcommand{\Etwo}{\mathrm{E2M1}}

\title{Integer-Exact Microscaling: Accurate 4-Bit\\Weight Formats for INT8 Matrix Units}

\ifpreprint
\author{Shinyoung Kang$^{1}$ \quad Junhyung Ahn$^{2}$ \quad Yujun Kim$^{3}$ \quad Jungho Park$^{4}$ \quad Andrew Jamieson$^{5}$ \\
$^{1}$UT Southwestern Medical Center \\
$^{2}$Department of Data and Decision Sciences, Emory University \\
$^{3}$Fuqua School of Business, Duke University \\
$^{4}$Heinz College of Information Systems and Public Policy, Carnegie Mellon University \\
$^{5}$Department of Bioinformatics, UT Southwestern Medical Center \\
\texttt{shinyoung.kang@utsouthwestern.edu}
}
\else
\author{Anonymous authors \\
Paper under double-blind review}
\fi

\begin{document}

\maketitle
\ifpreprint\lhead{Preprint}\fi

\begin{abstract}
4-bit weights make large language models fit on personal hardware, and the most accurate formats use fine-grained float block scales. Many matrix units, however, are fastest on INT8$\times$INT8, requiring per-block partial-sum scaling or FP16 conversion. We ask how much accuracy it costs to avoid this. Any format whose weights are one row scale times an integer block multiplier times an integer code runs as one exact INT8 GEMM with one rescale per output when the largest product is at most 127. A bound explains why integer multipliers track a block's ideal scale where powers of two cannot; in a 134-format sweep, doubled E2M1 codes with multipliers $1..10$, the INT8 limit, lie near the optimum. With INT8 activations and calibration-free rounding, 8-wide blocks (4.51 bits) beat NVFP4 (4.5 bits) with scale search on all eight models tested (four families, up to a 125B MoE), with paired intervals excluding zero on every one; paired with zero-overhead folded SmoothQuant, calibrated 8-wide beats calibrated NVFP4 and W4A16 NVFP4 on all three dense models tested. 16-wide blocks (4.26 bits) match standard NVFP4 on seven of eight models and solve 94.8\% of GSM8K on the 125B model (8-wide: 95.3\%, vs.\ 91.0\% for MLX's MXFP4). In MoE models, dense projections ($<4\%$ of weights) carry over half of the loss and decode bytes. Standard MXFP4 joins the family when a row's block exponents span at most three (97.9--99.97\% of rows across eight models). Exact W4A8 prefill runs $1.43$--$1.72\times$ faster than MLX on an M5 dense model and $1.12$--$1.60\times$ on three MoEs up to 125B (decode $0.98$--$1.03\times$ SC-MXFP4, $0.90$--$0.98\times$ 8-wide), and on an NVIDIA L4 prefills Qwen3-4B and Mistral-7B $1.18$--$1.37\times$ faster than FP16 ($1.22$--$1.68\times$ vs.\ Marlin W4A16) at $2.4$--$3.0\times$ lower peak VRAM.
\end{abstract}

\section{Introduction}

Running large language models (LLMs) on personal hardware depends on 4-bit weight quantization: memory is limited and
token generation is bound by memory bandwidth, so bytes per weight dominate. Work on 4-bit formats has therefore asked which
scales and codebooks lose the least accuracy~\citep{awq,qlora,mrgptq,fourover6}, and fine-grained float block scales usually win: NVFP4~\citep{nvfp4} is usually more accurate than MXFP4~\citep{ocpmx}. Hardware pulls the other way. Processing the prompt (prefill) is compute-bound, and many matrix units are faster on INT8 but have no FP4:
the INT8 tensor cores of NVIDIA GPUs before Blackwell~\citep{nvidia2020a100,nvidia2022ada}, AMD's~\citep{amd2023cdna3} and Intel's~\citep{intel2024amx} integer matrix units, Arm's SME~\citep{hellosme}, and the Neural
Accelerators of Apple's M5 GPUs~\citep{basertm5}, our case study, where INT8$\times$INT8 runs at about twice the FP16
rate but only when \emph{both} operands are integers. Deployed formats are instead dequantized to 16-bit floats inside the multiplication (W4A16)~\citep{mlx,llamacpp,marlin}, so prefill runs at the 16-bit rate.

\textbf{How much accuracy does it cost to make 4-bit weights exact INT8 operands, with no per-block work in the reduction?} At NVFP4's bit rate with INT8 activations and calibration-free rounding, none: an integer-exact format beats NVFP4 with scale search on all eight models (with block-Gram search~\citep{soscale,osfp4} alone, the two split 1--1; with zero-overhead folded SmoothQuant~\citep{smoothquant}, calibrated 8-wide beats calibrated NVFP4 and W4A16 NVFP4 on all three dense models tested). Against unfolded W4A16 NVFP4, plain INT8 costs up to 0.19 perplexity on four of eight models. One integer budget, $\max\mathcal{C}\cdot\max\mathcal{M}\le127$, leaves enough room for a good codebook and block multipliers fine enough to track each block's ideal scale.

\begin{figure*}[t]
  \centering
  \includegraphics[width=0.90\textwidth]{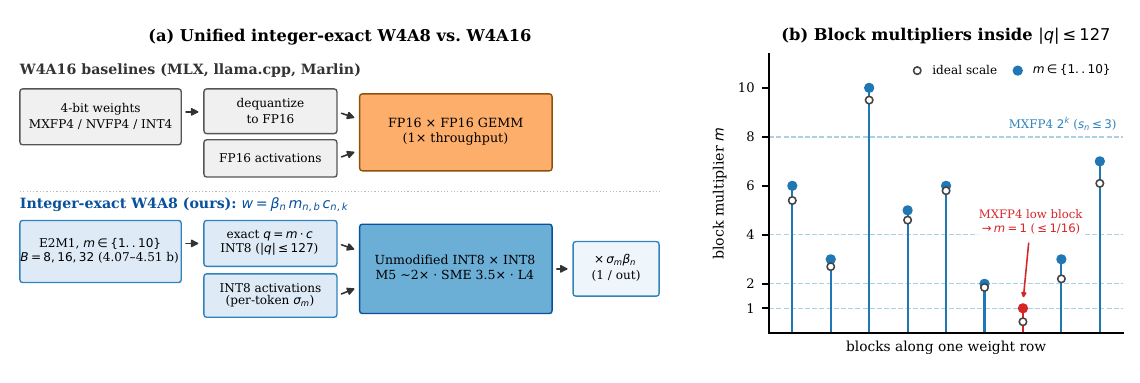}
  \vskip -0.10in
  \caption{(a) Unified integer-exact W4A8 vs.\ W4A16: weights $w=\beta_n m_{n,b} c_{n,k}$ with $|q|=|m_{n,b}c_{n,k}|\le127$ unpack to exact INT8 vectors paired with per-token INT8 activations on unmodified INT8 GEMMs across M5, Arm SME, and NVIDIA A100/L4 with one rescale per output. (b) Integer block multipliers $m\in\{1..10\}$ track ideal block scales closely across a row (Proposition~\ref{prop:ladder}); standard MXFP4 is the $m\in\{1,2,4,8\}$ ($s_n\le3$) special case, clamping low blocks to $m=1$ at error $\le1/16$ (Proposition~\ref{prop:cost}).}
  \label{fig:overview}
  \vskip -0.22in
\end{figure*}

Integer scales are not new: QServe~\citep{qserve} and DGQ~\citep{dgq} pair INT4 codes with integer group scales, while llama.cpp, vLLM, and concurrent M5 kernels~\citep{llamacppfp4,mlspecqd,vllmw4a8mxfp4,cider,omlx,scratchy,omlxmx} apply per-group partial-sum scaling. Folding scales into INT8 operands matches rather than beats per-group M5 kernels in speed (\S\ref{sec:q38}); it adds \textbf{portability, exactness, and format accuracy}. Running unmodified INT8 GEMMs transfers across Apple M5, Arm SME, and NVIDIA A100/L4, while E2M1 cuts Qwen3-1.7B perplexity from 12.10 (affine INT4) to 10.27 ($2.4\times$ lower output error, $5.99$ vs.\ $14.27\times10^{-3}$), cuts 16-wide error vs.\ INT4 ($7.36$ vs.\ $12.38\times10^{-3}$; Table~\ref{tab:formats}), cuts 125B FP8 loss $3\times$ vs.\ OCP MXFP4 ($+3.9\%$ vs.\ $+11.8\%$, $+4.3$ GSM8K pts), and preserves MXFP4 files for SC-MXFP4 (Figure~\ref{fig:overview}).

\paragraph{Contributions.}
\begin{itemize}
  \item \textbf{An integer-exact FP4 format family and its design space} (\S\ref{sec:family}, \S\ref{sec:space}): the INT8-closed
        instance of two-level integer scaling~\citep{vsquant,iq4ks,dgq,qserve} with the FP4 codebook, a bound
        that explains why integer multipliers beat powers of two, and a sweep of 134 formats. 8-wide
        blocks beat NVFP4 with scale search at its bit rate; 16-wide blocks match standard NVFP4 at 0.24 fewer bits on seven of eight models.
  \item \textbf{Standard MXFP4 as a member} (\S\ref{sec:mxfp4}): a row of MXFP4 weights equals a
        symmetric INT8 vector times $2^{E-1}$, with $E$ its smallest block exponent, if and only if its
        block exponents span at most three (when the top block reaches 4 or 6, as under the OCP rule), with a constraint-enforcing quantizer, cost bound, and row-spread null model extending~\citet{adamx} to an exactness condition.
  \item \textbf{Evidence across eight models up to 125B} (\S\ref{sec:acc}, \S\ref{sec:q38}): perplexity and zero-shot accuracy, 97.9--99.97\% compliance, SmoothQuant beating calibrated/W4A16 NVFP4 (Table~\ref{tab:k1_sq}), MoE loss/decode breakdowns, INT8 acts $\le 0.3\%$ at 125B, and a rotation bound (App.~\ref{app:ablation}).
  \item \textbf{Complete, exact W4A8 systems} (\S\ref{sec:system}, \S\ref{sec:speed}): faster M5 prefill on a dense model and three MoEs up to 125B ($0.98$--$1.03\times$ MLX decode for SC-MXFP4, $0.90$--$0.98\times$ for 8-wide; Table~\ref{tab:q38speed}), bit-identical decode logits, Arm SME GEMMs, and NVIDIA L4 prefill beating FP16/Marlin~\citep{marlin} (plus A100 layers; Tables~\ref{tab:nvidia}--\ref{tab:nvidia_e2e}).
\end{itemize}

\section{Background and motivation}
\label{sec:char}

\paragraph{MXFP4.} An MXFP4 tensor stores each weight as $w = 2^{e_b}\,v$ with $v \in \Etwo =
\{0,\pm\tfrac12,\pm1,\pm\tfrac32,\pm2,\pm3,\pm4,\pm6\}$ and one 8-bit exponent (E8M0) $e_b$ per block $b$ of 32
consecutive values along the reduction dimension, 4.25 bits per weight~\citep{ocpmx,rouhani2023microscaling}. NVFP4 uses 16-value
blocks with FP8 (E4M3) scales~\citep{nvfp4} and is usually more accurate, but its scales are not powers of two.

\paragraph{The M5 Neural Accelerators.} The M5 GPU's matrix units are programmed through Metal~4 tensors and
the \texttt{matmul2d} operation~\citep{wwdcmetaltensors}. The shipped headers (macOS~26.6) accept
\texttt{half}, \texttt{bfloat}, \texttt{float}, \texttt{int8} and \texttt{uint8} operands and a packed 4-bit
integer type, but no FP8 or FP4 type; macOS~27.2 adds FP8, FP4, and block-scaled MXFP4 operands, but none exceeds the FP16 rate (FP8 runs at $0.97$--$1.00\times$ FP16, FP4 at $0.78$--$0.85\times$, and block-scaled MXFP4 at $0.27$--$0.46\times$, while INT8$\times$INT8 stays at $1.91$--$1.97\times$; Appendix~\ref{app:micro}), and Apple does not publish
per-type throughput. We measured it with
minimal \texttt{matmul2d} kernels whose inputs are small integers, so that every operand type computes the
same product (verified bit-exactly), timing all types interleaved after a clock warm-up
(Appendix~\ref{app:micro}): INT8$\times$INT8 reaches 109--127 TOPS (Figure~\ref{fig:micro}),
$1.79$--$1.98\times$ the FP16 rate, while FP16$\times$INT8 and FP16$\times$INT4 run at $0.68$--$0.96\times$
FP16, because the hardware converts the integer operand. The native INT8$\times$INT4 path reaches
$1.70$--$1.82\times$ FP16, but its 4-bit operand holds $-8\ldots7$ and cannot represent MXFP4 values.
The integer rate therefore requires integer activations \emph{and} integer weights. Similar ratios have since been reported on other M5 and M6 parts~\citep{mlxissue4627,omlx,scratchy}.

\section{Integer-exact formats}
\label{sec:method}

\subsection{The format family}
\label{sec:family}

A linear layer runs as one INT8 GEMM with no per-block work if each weight row is one scale times an INT8 vector. That holds for any format of the form
\begin{equation}
  w_{n,k} = \beta_n \, m_{n,b(k)} \, c_{n,k}, \quad c \in \pm\mathcal{C},\; m \in \mathcal{M} \subset \{1,2,\ldots\},
  \label{eq:family}
\end{equation}
with an integer magnitude codebook $\mathcal{C}$, a set $\mathcal{M}$ of small integer block multipliers and one float
scale $\beta_n$ per row, provided $\max\mathcal{C} \cdot \max\mathcal{M} \le 127$. The operands $q = m\,c$ are then symmetric INT8, and with per-token INT8 activations $a_{m,k}$ and scales $\sigma_m$,
\begin{equation}
  y_{m,n} = \sigma_m\, \beta_n \textstyle\sum_k a_{m,k}\, q_{n,k},
\end{equation}
with one rescale per output; INT32 accumulation cannot overflow for $K \le (2^{31}-1)/127^2 \approx 133$k. The two features that make
NVFP4 more accurate than MXFP4, 16-wide blocks and block scales finer than powers of two, are both available inside
this constraint: with doubled E2M1 elements, $\mathcal{M} = \{1,\ldots,10\}$ gives ten block-scale levels per row.
We quantize by trying, per row, eight values per octave below the largest $\beta_n$ that represents the row
maximum, $\beta_{\max}2^{-j/8}$ with $\beta_{\max} = \max_k|w_{n,k}|/(\max\mathcal{C}\,\max\mathcal{M})$ and
$j = 0,\ldots,7$, taking per block the multiplier with the least squared error and per row the $\beta_n$ with the
least total error. Storage is $4 + \lceil\log_2|\mathcal{M}|\rceil/B + 32/K$ bits per weight for block size $B$. These formats are not MXFP4 files: they
trade compatibility for accuracy.

\subsection{Why these multipliers: the design space}
\label{sec:space}

Since $\max\mathcal{C}\cdot\max\mathcal{M}\le127$, every format spends one integer budget on the codebook's shape and on the
range of block scales. Doubled E2M1 needs integers up to 12, its smallest integer form, so $m\le\lfloor127/12\rfloor=10$ is
the limit for that shape, not a tuned value. The budget buys scale accuracy where it matters:

\begin{proposition}[Multiplier ladder]
\label{prop:ladder}
Let a block have peak $a>0$ in a row with peak $A$, let $\mathcal{M}=\{1,\ldots,M\}$, $C=\max\mathcal{C}$ and $\beta=A/(CM)$. Then
$m=\lceil a/(\beta C)\rceil\in\mathcal{M}$ does not clip the block and its scale $s=\beta m$ satisfies
$a/C\le s<(a/C)(1+A/(Ma))$. With largest codebook gap $g$, every element then errs by at most
$gs/2<\tfrac{g}{2C}(a+A/M)$. With $\mathcal{M}=\{1,2,4,8\}$ the smallest non-clipping scale can be almost $2a/C$ for any $a$.
\end{proposition}

\begin{figure}[t]
  \centering
  \includegraphics[width=0.65\textwidth]{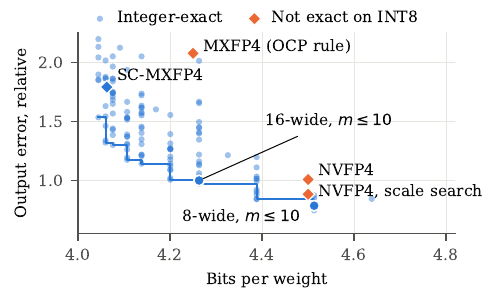}
  \vskip -0.1in
  \caption{Layer-output error vs.\ bits per weight, geometric mean over Qwen3-1.7B, Qwen3-4B and SmolLM3, relative to the 16-wide format with $m\le10$. The
  line is the best integer-exact format at each bit rate; points above $2.25$ are not shown.}
  \label{fig:space}
  \vskip -0.1in
\end{figure}

The excess over the free scale $a/C$ is therefore additive for integer multipliers, $\beta$ regardless of $a$, and smallest in
relative terms for the largest blocks, which dominate the error, while powers of two can double the scale of any block. The proof is in Appendix~\ref{app:proofs}. On Qwen3-1.7B's real block peaks, the energy-weighted squared scale excess,
$\sum s^2/\sum(a/C)^2$, is $1.18$ for $m\le10$ and $2.15$ for powers of two.
The measured error confirms it (Figure~\ref{fig:space}; 134 formats, three models, every third layer; Appendix~\ref{app:space}): relative to a free
float scale per block, weight error falls from $1.8\times$ at $M=1$ to $1.00\times$ at $M=10$ on all three models,
and layer-output error to $1.00$--$1.08\times$ on two of them and $1.36\times$ on Qwen3-1.7B, where $94.3\%$ of the excess comes from \texttt{mlp.down\_proj} activation spikes (chiefly layer~27) and collapses to $0.11\times$ under block-Gram weighting (Appendix~\ref{app:ablation}). Other
codebook shapes do not consistently win: at 16-wide blocks the best alternatives have $2$--$3\%$ less output error on
average but change from model to model, and one of them is much worse in perplexity (\S\ref{sec:acc}). Block
width trades bits for accuracy: $B=8$ costs 0.25 bits per weight over $B=16$ and removes $21\%$ of its output error (geometric mean).

\subsection{Standard MXFP4 as a member}
\label{sec:mxfp4}

Doubling an E2M1 value gives an integer, $2v \in \{0,\pm1,\pm2,\pm3,\pm4,\pm6,\pm8,\pm12\}$, a fact existing
MXFP4 kernels use to form INT8 operands~\citep{llamacppfp4,mlspecqd}. For a weight
row $n$ with block exponents $e_{n,b}$, let $E_n = \min_b e_{n,b}$, the \emph{spread}
$s_n = \max_b e_{n,b} - E_n$, and
\begin{equation}
  q_{n,k} = 2v_{n,k}\cdot 2^{\,e_{n,b(k)} - E_n}, \qquad w_{n,k} = 2^{E_n-1}\, q_{n,k}.
  \label{eq:embed}
\end{equation}

\begin{proposition}[Exact embedding]
\label{prop:embed}
$q_{n,k}$ is an integer for every $k$ and $2^{E_n-1}q_{n,\cdot}$ reproduces the MXFP4 row exactly. If
$s_n \le 3$, then $|q_{n,k}| \le 96$, so the row is a symmetric INT8 vector ($|q|\le127$) times one scale. Conversely, if
$s_n \ge 4$ and a block at the largest exponent contains $\pm4$ or $\pm6$ (as every block whose exponent follows the
OCP rule does), some $|q_{n,k}| \ge 128$.
\end{proposition}

The proof is in Appendix~\ref{app:proofs}. Such a row is the instance $\mathcal{C} = \{0,1,2,3,4,6,8,12\}$, $\mathcal{M} = \{1,2,4,8\}$, 32-wide blocks and $\beta_n=2^{E_n-1}$ of Eq.~\ref{eq:family}, and INT32 accumulation cannot overflow for $K \le (2^{31}-1)/(127\cdot 96) \approx 176$k.

\paragraph{Spread-constrained quantization.} We quantize to MXFP4 by choosing each block exponent by squared-error search over the OCP rule
$\lfloor\log_2 \max_k|w_k|\rfloor - 2$ and its two neighbors, then raising any exponent below
$\max_b e_{n,b}-3$ to that value. We call the result \emph{spread-constrained MXFP4} (SC-MXFP4). It is a valid
MXFP4 tensor that any MXFP4 decoder reads unchanged, and by Proposition~\ref{prop:embed} it runs exactly on
INT8 units. Raising an exponent only coarsens blocks whose values are small relative to the row:

\begin{proposition}[Cost of the constraint]
\label{prop:cost}
Let $E^* = \max_b e_{n,b}$. Every element of a block whose exponent is raised to $E^*-3$ has absolute rounding
error at most $2^{E^*-4}$, which is $1/16$ of $2^{E^*}$, the largest error of round-to-nearest between the grid
points of a block at exponent $E^*$.
\end{proposition}

If exponent search gives the top block the exponent above its OCP value, that block's own values stay below
$4\cdot2^{E^*}$ and err by at most $2^{E^*-1}$, so the coarsened blocks then err by at most $1/8$ of the top block's
actual worst case.

\paragraph{Why real weights satisfy the condition.}
\label{sec:why}
Under the OCP rule, $s_n \le 3$ whenever the largest block maximum of a row is less than $8\times$ its smallest
(and only if it is less than $16\times$), so the condition is about the dynamic range of block maxima within a row. For rows of i.i.d.\ Gaussian weights,
maxima of 32 values concentrate tightly: over 50k simulated rows per width ($K = 1024$--$6144$; unit scale, exponents chosen by
the same search as for real weights), the median of
$\log_2(\max_b/\min_b)$ is 1.1--1.5, its 99.99th percentile 2.0--2.3, and no row violates the condition. Rows of
heavy-tailed i.i.d.\ weights (Student-$t$, 4 degrees of freedom) violate it in 2.7--37\% of cases. Real weight
rows therefore behave like light-tailed rows \emph{within each row}, even though weight distributions pooled across
rows are heavy-tailed; the few violations are dead channels and outlier blocks (Appendix~\ref{app:spread}), and the blocks
that the constraint coarsens are the small ones, which contribute little to the output (\S\ref{sec:acc}).

\paragraph{Composing with GPTQ.} Because the exponents are fixed before rounding, GPTQ~\citep{gptq} can choose each
element's E2M1 value with them held fixed; the result is still a standard MXFP4 file and still exact on INT8 units.

\section{Accuracy}
\label{sec:acc}

The kernels are exact, so accuracy depends only on the quantization scheme: the format's weights and symmetric
per-token INT8 activations in every transformer-block linear layer, with embeddings, norms, attention, routers
and the output head in BF16. We evaluate it with fake quantization, which matches the deployed kernels to within
0.005 perplexity on the full test set (Qwen3-1.7B and Qwen3.5-35B-A3B, three formats each, \S\ref{sec:speed}; within 0.011 on Qwen3-4B and Mistral-7B; within 0.4\% on 8 windows of the 125B model, \S\ref{sec:q38}), on Qwen3-0.6B/1.7B/4B~\citep{qwen3}, SmolLM3-3B~\citep{smollm3}, Mistral-7B-v0.3~\citep{mistral7b}, the
MoE OLMoE-1B-7B~\citep{olmoe} and the MoE Qwen3.5-35B-A3B (WikiText-2~\citep{wikitext} test perplexity, 2048-token
windows, full test set).

\begin{table*}[t]
  \centering\small
  \caption{WikiText-2 perplexity. W4A16: 4-bit weights, 16-bit activations; W4A8: per-token INT8 activations.
  SC-MXFP4 W4A8 is what our kernels compute; SC-MXFP4 packs at 4.07 bits via 2-bit shifts (\S\ref{sec:system}) or stores losslessly as a 4.25-bit MXFP4 file. All methods quantize the same matrices. $^\ddagger$MoE: every expert
  is quantized, the router stays BF16. Bits at $K = 4096$ (the FP32 row scales of our formats add $32/K$). -- = not
run.}
  \label{tab:ppl}
  \setlength{\tabcolsep}{3pt}\resizebox{\textwidth}{!}{\begin{tabular}{lccrrrrrrr}
\toprule
 & & & \multicolumn{3}{c}{Qwen3} & SmolLM3 & Mistral & OLMoE & Qwen3.5 \\
\cmidrule(lr){4-6}
Method & Bits & INT8 & 0.6B & 1.7B & 4B & 3B & 7B & 1B-7B$^\ddagger$ & 35B-A3B$^\ddagger$ \\
\midrule
BF16 & 16 &  & 12.67 & 9.40 & 7.90 & 7.91 & 5.32 & 7.49 & 5.72 \\
\midrule
INT4 affine g64 (MLX) W4A16 & 4.50 & no & 15.83 & 12.10 & 8.56 & 8.44 & 5.47 & 7.86 & 5.92 \\
NVFP4 W4A16 & 4.50 & no & 14.44 & 10.23 & 8.26 & 8.29 & 5.45 & 7.82 & 5.87 \\
NVFP4 W4A16, scale search & 4.50 & no & -- & 10.08 & -- & -- & -- & -- & -- \\
NVFP4 W4A8 & 4.50 & no & 14.91 & 10.51 & 8.47 & 8.36 & 5.51 & 7.92 & 5.93 \\
NVFP4 W4A8, scale search & 4.50 & no & 14.63 & 10.34 & 8.44 & 8.30 & 5.49 & 7.84 & 5.91 \\
\midrule
\textbf{SC-MXFP4 W4A8} (MXFP4-compat.) & 4.07 & yes & 15.55 & 10.99 & 8.75 & 8.62 & 5.59 & 8.11 & 6.05 \\
\quad + GPTQ & 4.07 & yes & -- & 10.52 & -- & -- & -- & -- & -- \\
\textbf{E2M1, 32-blocks, $m\le8$ W4A8} & 4.10 & yes & 14.94 & 10.86 & 8.57 & 8.41 & 5.53 & -- & -- \\
\textbf{E2M1, 16-blocks, $m\le10$ W4A8} & 4.26 & yes & 14.59 & 10.39 & 8.42 & 8.35 & 5.52 & 7.85 & 5.91 \\
\textbf{E2M1, 8-blocks, $m\le10$ W4A8} & 4.51 & yes & \textbf{14.19} & \textbf{10.27} & \textbf{8.37} & \textbf{8.25} & \textbf{5.48} & \textbf{7.77} & \textbf{5.89} \\
\bottomrule
\end{tabular}
}
\end{table*}

\paragraph{Integer-exact formats reach and pass NVFP4.} In perplexity with INT8 activations (Table~\ref{tab:ppl}), the 8-wide
format (4.51 bits) beats NVFP4 with scale search on all eight models (four families): by $0.44$, $0.07$, $0.07$ and $0.05$ on the four small models, by $0.013$ on Mistral-7B-v0.3 (5.480 vs.\ 5.492), by $0.070$ on OLMoE-1B-7B (7.769 vs.\ 7.839), by $0.020$ on Qwen3.5-35B-A3B (5.887 vs.\ 5.907), and by $0.075$ on Qwen3.8-Flash-Next (4.842 vs.\ 4.917, $[-0.095,-0.056]$, \S\ref{sec:q38}), with 95\% paired bootstrap intervals excluding zero on every model (Appendix~\ref{app:ci}; Mistral: $[-0.019,-0.006]$; 35B: $[-0.029,-0.012]$). The 16-wide format (4.26 bits) beats standard NVFP4 on
Qwen3-0.6B, 1.7B, 4B and OLMoE (tied on SmolLM3 and Mistral-7B), at 0.24 fewer bits; against NVFP4 with scale search it is ahead on Qwen3-0.6B, tied on Qwen3-4B and Qwen3.5-35B-A3B, and behind elsewhere. Non-E2M1 codebooks are not a safe replacement: at 16-wide blocks $\{0,1,2,3,4,6,8,10\}$ is
$0.04$--$0.38$ worse on two models and better on one, $0.01$ on the last; at 8-wide blocks with $m\le12$ it is ahead of E2M1 on three of the four small models (by $0.02$--$0.18$) and tied on Qwen3-4B (Table~\ref{tab:ci}). With uniform INT4 elements, the structure of QServe-style schemes, layer-output error at 16-wide blocks is $1.73\times$ higher on Qwen3-1.7B but within 4\% on Qwen3-4B and SmolLM3 (Table~\ref{tab:formats} in Appendix~\ref{app:space}), so E2M1's non-uniform grid matters on some models, not all. Wider calibration-free NVFP4 searches (Four-Over-Six~\citep{fourover6}, $\pm 6$ E4M3) still trail 8-wide on Qwen3-1.7B and 4B (Table~\ref{tab:nvfp4_search}, Appendix~\ref{app:ablation}); block-Gram search~\citep{soscale,osfp4} (where multi-octave $\beta_n$ saturates at 1 octave) splits 1--1 (1.7B: $-0.093$; 4B: $+0.027$). Zero-overhead folded SmoothQuant~\citep{smoothquant} cuts W16A8 error, flips calibrated 8-wide ahead of calibrated NVFP4 on all three dense models ($-0.011$ to $-0.049$) and W4A16 NVFP4 ($-0.095$ to $-0.470$), and puts Mistral-7B RTN 8-wide at 5.430 vs.\ 5.439 (SmoothQuant NVFP4 search) and 5.451 (W4A16 NVFP4; Table~\ref{tab:k1_sq}).

\paragraph{Downstream tasks on the deployed system.} Table~\ref{tab:tasks} reports zero-shot accuracy
(lm-evaluation-harness), every W4A8 pass running through our kernels, next to MLX's MXFP4 W4A16 on the same matrices (MLX's default conversions also quantize embeddings and head and score 0.3--1.2 points lower). SC-MXFP4 W4A8 matches it: 61.9 vs.\ 61.8 on Qwen3-1.7B and 75.5 vs.\ 75.7 on Qwen3.5-35B-A3B (MLX's default affine 4-bit conversion: 75.8). On Qwen3-1.7B the 8-wide format scores 63.5 (16-wide: 62.7), 1.0 below BF16 and 1.6 $[1.0, 2.3]$ above SC-MXFP4 (paired bootstrap 95\% interval). On Qwen3.5-35B-A3B it is the lowest 4-bit system (74.8--75.7), 0.7 $[-1.4, -0.1]$ below SC-MXFP4, mostly on ARC, although its log-likelihoods and KL divergence are the closest to BF16 (Appendix~\ref{app:kl}): there its advantage is fidelity, not task accuracy.

\begin{table}[t]
\centering
\small
\caption{Zero-shot accuracy (\%) of the deployed systems (Qwen3.5: first 1000 examples per task). MXFP4 and
W4A8 quantize the same matrices; ``default'' is MLX's conversion, which also quantizes embeddings and head.
Normalized accuracy for ARC-c, HellaSwag and PIQA.}
\label{tab:tasks}
\setlength{\tabcolsep}{3.5pt}
\resizebox{\linewidth}{!}{%
\begin{tabular}{llrrrrrrr}
\toprule
Model & Method & ARC-e & ARC-c & HS & PIQA & WG & LAMB. & Mean \\
\midrule
Qwen3-1.7B   & BF16                    & 73.4 & 45.0 & 66.4 & 75.5 & 63.6 & 63.2 & 64.5 \\
             & MXFP4 W4A16 (default)   & 70.5 & 43.5 & 62.9 & 74.6 & 63.3 & 49.0 & 60.6 \\
             & MXFP4 W4A16             & 72.5 & 44.7 & 63.9 & 74.3 & 62.8 & 52.3 & 61.8 \\
             & W4A8, SC-MXFP4 (ours)   & 73.2 & 43.6 & 63.5 & 73.6 & 62.3 & 55.2 & 61.9 \\
             & W4A8, 16-wide (ours) & 72.9 & 43.8 & 64.9 & 74.0 & 62.9 & 57.6 & 62.7 \\
             & \textbf{W4A8, 8-wide (ours)} & 73.0 & 44.8 & 65.2 & 74.6 & 63.4 & 60.2 & \textbf{63.5} \\
\midrule
Qwen3.5-35B-A3B & BF16                 & 85.0 & 62.4 & 75.7 & 84.0 & 76.3 & 75.6 & 76.5 \\
             & MXFP4 W4A16 (default)   & 83.2 & 62.8 & 75.1 & 83.3 & 75.6 & 72.3 & 75.4 \\
             & MXFP4 W4A16             & 84.6 & 63.2 & 75.4 & 83.6 & 74.6 & 73.0 & 75.7 \\
             & W4A8, SC-MXFP4 (ours)   & 84.3 & 63.8 & 75.1 & 83.4 & 73.8 & 72.7 & 75.5 \\
             & W4A8, 16-wide (ours)   & 83.9 & 61.9 & 75.5 & 84.1 & 73.4 & 72.2 & 75.2 \\
             & W4A8, 8-wide (ours)    & 83.4 & 61.0 & 75.3 & 83.5 & 73.2 & 72.4 & 74.8 \\
\bottomrule
\end{tabular}}
\end{table}

\paragraph{GPTQ and rotations.} In simulation on three models (Appendix~\ref{app:ablation}), GPTQ~\citep{gptq}, calibrated on WikiText-2 \emph{train}, lowers W4A8 perplexity of every format, and the 8-wide format stays ahead of NVFP4 with scale search. Adding Hadamard input rotations brings it within 0.13--0.22 of BF16, partly by shrinking the INT8 step (Proposition~\ref{prop:rot}). Rotation precedes the INT8 step, so the GEMM stays exact; in our Qwen3.5-35B-A3B kernels (round-to-nearest weights) it lowers perplexity from 5.889 to 5.849 at a cost of 1--4\% in prefill speed (separate processes). On the kernels GPTQ lowers SC-MXFP4's zero-shot mean from 61.9 to 60.6, so round-to-nearest stays our main configuration.

\paragraph{The spread constraint is free.} SC-MXFP4 and unconstrained MXFP4 (both with exponent search, W4A16) differ by
at most 0.007 perplexity on four models, and the relative weight error by at most $0.01\times10^{-3}$ on all eight (Appendices~\ref{app:ablation}, \ref{app:spread}).

\paragraph{INT8 activations are cheap.} Per-token INT8 activations add $+0.07$--$0.36$ perplexity over W4A16 ($+0.8\%$ on Qwen3.5-35B-A3B), matching the BF16$\to$W16A8 gap ($+0.09$--$0.32$), so the errors barely compound. Most arises at the down-projection input; keeping each token's 4 largest inputs in float removes 82--85\% of the loss without changing weights, lowering 8-wide perplexity from 10.27 to 10.07 on Qwen3-1.7B and from 8.37 to 8.20 on 4B without calibration (Table~\ref{tab:bypass}).

\paragraph{Where an MoE's loss sits.} On Qwen3.5-35B-A3B the losses of the two classes of matrices add (Table~\ref{tab:scope}): over BF16 weights with INT8 activations, 8-wide experts cost $+0.048$ and 8-wide dense projections (DeltaNet, attention) $+0.060$, although the dense projections hold 3.8\% of the quantized weights; they are, however, half of the bytes a decode step reads. SC-MXFP4 experts with 8-bit dense projections match the 8-wide format everywhere at 4.24 bits but $1.37\times$ the decode bytes (simulated; 125B in \S\ref{sec:q38}).

\paragraph{SC-MXFP4 against other 4-bit formats.} NVFP4 and, on four of seven models, MLX's affine
INT4 (group 64) are 0.1--0.6 perplexity more accurate than SC-MXFP4, and llama.cpp's Q4\_K\_M (about 5 bits) more accurate than all our formats (Appendix~\ref{app:llamacpp}); none of them runs exactly on INT8 units.

\section{A W4A8 system on Apple M5}
\label{sec:system}

\paragraph{Weight layout.} We store a lossless re-layout of SC-MXFP4: a 4-bit code per weight (sign and E2M1
index), the block shift $e_{n,b} - E_n \in \{0,\ldots,3\}$ in 2 bits per 32 weights, and one FP32 scale per
row: $4 + 2/32 + 32/K$ bits per weight for reduction dimension $K$ (4.07 at $K = 4096$, 4.11 for the $K = 640$
expert down projections of \S\ref{sec:q38}; MXFP4: 4.25). Decoding is $q = \pm\,\mathrm{LUT}[\mathrm{idx}] \ll
\mathrm{shift}$ with $\mathrm{LUT} = (0,1,2,3,4,6,8,12)$, packed as the nibbles of one 32-bit constant. Only the E8M0 scales are re-encoded, offline, and each byte is recovered exactly, so SC-MXFP4 can also
be stored, and is read by MLX's kernels, as a standard 4.25-bit MXFP4 tensor.

\paragraph{Dense prefill.} Every token multiplies the same weights, so each layer's weights are unpacked once
per call into an INT8 buffer (a temporary of $N\cdot K$ bytes that MLX's allocator recycles across layers) and
multiplied by a plain INT8 \texttt{matmul2d} reading device memory. Unpacking
inside the GEMM's K-loop instead, as W4A8 kernels on NVIDIA GPUs do~\citep{qserve}, re-unpacks the same tile
for every row tile and runs at $0.39$--$0.48\times$ FP16 on M5 (Appendix~\ref{app:kernels}). Activations are
quantized per token, fused with the preceding RMSNorm; q/k/v and gate/up each run as one GEMM on stacked
weights, with gate and up rows interleaved in pairs so that SwiGLU's two inputs share a register tile; the scales,
residual additions and SwiGLU are applied in the GEMM's registers. Every non-GEMM step
reproduces MLX's rounding, so the model's logits are bit-identical to running each layer separately.

\paragraph{MoE prefill.} One counting-sort kernel orders the token--expert assignments by expert, the INT8
activation rows are gathered in that order, and all experts run in one grouped INT8 GEMM over their contiguous rows;
a combine kernel reads each token's expert outputs back by position and weights them by the routing scores. Short prompts unpack each weight tile into threadgroup memory inside the kernel; once the average
expert, the shared one included, has 256 rows ($L(k{+}1)/(E{+}1) \ge 256$), the two-pass scheme above takes over.

\paragraph{Decode.} A GEMV reads the packed weights directly and computes the same integer accumulations;
decode logits are bit-identical to running the prefill kernels one token at a time (checked on Qwen3-1.7B, in the three
formats, on OLMoE-1B-7B and on Qwen3.5-35B-A3B).

\section{Speed}
\label{sec:speed}

All timings are on an M5 Max (128\,GB, AC power, High Power mode; macOS~26.6, and 27.2 where marked), on an otherwise idle GPU, with
methods interleaved in one process (125B: one system per process) and medians reported; for Qwen3-1.7B and Qwen3.5-35B-A3B, MLX MXFP4's times are checked against reference values (Appendix~\ref{app:method}). Baselines are MLX~0.32 with its MXFP4 and affine 4-bit checkpoints and llama.cpp
(commit f74e2fa, built from source, Q4\_0 and MXFP4\_MOE; Q4\_K\_M and IQ4\_XS in Appendix~\ref{app:llamacpp}). Both baselines already use the M5 matrix units, with FP16
operands. In decode, our system uses MLX's MXFP4 output projection, as MLX's own checkpoint does, so decode timings compare the transformer blocks; the accuracy results use a BF16 output head.

\paragraph{Linear layers.} On LLM-sized layers (512--4096 tokens, $N,K$ up to 14336) W4A8, including activation quantization, unpack
and scaling, is $1.37$--$1.90\times$ faster than MLX's MXFP4 \texttt{quantized\_matmul} and $1.48$--$1.64\times$ faster than the identical two-pass design with FP16 operands,
which isolates the operand type (Appendix~\ref{app:layer}). On NVIDIA A100 and L4 Tensor Cores (same six shapes; Appendix~\ref{app:micro}), our layers stay bit-exact and beat Marlin W4A16~\citep{marlin} on L4 ($1.00$--$1.48\times$; $1.11$--$1.22\times$ vs.\ FP16) and at $M\ge1024$ on A100 ($1.05$--$1.44\times$). End to end on L4 (Table~\ref{tab:nvidia_e2e}), W4A8 matches simulated perplexity within $0.011$ at $2.4$--$3.0\times$ less VRAM than FP16, beating FP16 and Marlin on Mistral-7B ($1.18$--$1.35\times$ and $1.22$--$1.68\times$; $5.1$ vs.\ $14.0$\,GiB) and Qwen3-4B at $L\ge1024$ ($1.18$--$1.37\times$ and $1.29$--$1.46\times$; $0.90\times$/$0.86\times$ at $L=512$).

\paragraph{End to end.} Table~\ref{tab:system} reports time to first token (the whole prompt in one forward pass)
and decode throughput. On the dense Qwen3-1.7B, prefill is $1.43$--$1.72\times$ faster than MLX MXFP4 over both
clock regimes (Table~\ref{tab:regimes}) and $1.70$--$2.44\times$ faster than llama.cpp. On the MoEs the gain grows with prompt length, because each expert sees
only about $8L/E$ of an $L$-token prompt's rows and short prompts are bound by streaming expert weights for every
method: $1.21$--$1.60\times$ on OLMoE-1B-7B (64 experts) and $1.14$--$1.36\times$ on Qwen3.5-35B-A3B (256 experts; 8-wide format $1.12$--$1.34\times$; at 512 tokens only in a session of its own, see Table~\ref{tab:system}),
where W4A8 covers 99.93\% of the block-linear parameters and 86--96\% of prefill FLOPs. The packed weights equal the
fake-quantized reference for every matrix of all three models, every integer accumulation we checked is exact,
and WikiText-2 perplexity on the kernels equals the reference (Qwen3.5: 6.052 vs.\ 6.051; 16-wide 5.909 vs.\ 5.909). Decode runs from the same
4.1-bit weights at $0.98\times$ MLX's speed on Qwen3-1.7B and OLMoE and $1.01\times$ on Qwen3.5-35B-A3B and is $1.12$--$1.24\times$ faster than llama.cpp; with the 8-wide format it runs at $0.90$--$0.91\times$ on Qwen3.5-35B-A3B (both formats in one process) and $0.93\times$ on Qwen3-1.7B.
\begin{table}[t]
\centering
\small
\caption{Time to first token (ms) and decode (tokens/s), M5 Max. Speedups (unrounded medians): W4A8 vs.\ MLX MXFP4 in the same session, and vs.\ llama.cpp (Q4\_0, its fastest 4-bit type; separate session; Qwen3.5 on macOS~26.6.2). Prefill: sustained regime. Qwen3-1.7B and OLMoE (macOS~27.2): mean of two sessions (1.7B decode: one session). Qwen3.5 (macOS~27.2, three methods interleaved, anchor-checked): decode from one session with both formats; prefill at 2048/8192 is the mean of two such sessions ($1.08$ and $1.21\times$ at 2048), and 512 is a dedicated burst session (multi-length session details in Appendix~\ref{app:method}). $^\dagger$From a session whose MLX time failed the anchor check.}
\label{tab:system}
\setlength{\tabcolsep}{3.7pt}
\resizebox{\columnwidth}{!}{%
\begin{tabular}{llrrrrr}
\toprule
Model & Tokens & llama.cpp & MLX MXFP4 & \textbf{W4A8, SC-MXFP4} & vs.\ MLX & vs.\ llama.cpp \\
\midrule
Qwen3-1.7B (dense)        & 512  & 40.6   & 34.0   & \textbf{23.8}   & $1.43\times$ & $1.70\times$ \\
                          & 2048 & 171.8  & 127.4  & \textbf{74.6}   & $1.71\times$ & $2.30\times$ \\
                          & 8192 & 992.4  & 682.3  & \textbf{406.0}  & $1.68\times$ & $2.44\times$ \\
OLMoE-1B-7B (MoE)         & 512  & 41.6   & 39.6   & \textbf{31.8}   & $1.25\times$ & $1.31\times$ \\
                          & 2048 & 157.8  & 115.9  & \textbf{84.3}   & $1.37\times$ & $1.87\times$ \\
                          & 4096 & 336.8  & 234.9  & \textbf{162.7}  & $1.44\times$ & $2.07\times$ \\
Qwen3.5-35B-A3B (MoE)     & 512  & 140.3  & 113.0  & \textbf{97.7}   & $1.16\times$ & $1.44\times$ \\
                          & 2048 & 511.3  & 380.0  & \textbf{332.6}  & $1.14\times$ & $1.54\times$ \\
                          & 8192 & 2388.2 & 1530.0 & \textbf{1127.9} & $1.36\times$ & $2.12\times$ \\
\midrule
\multicolumn{7}{l}{\emph{Decode, 512-token context (tokens/s)}} \\
Qwen3-1.7B                &      & 290 & 366 & \textbf{361} & $0.98\times$ & $1.24\times$ \\
OLMoE-1B-7B               &      & 401 & 458 & \textbf{448} & $0.98\times$ & $1.12\times$ \\
Qwen3.5-35B-A3B           &      & 137$^\dagger$ & 151 & \textbf{153} & $1.01\times$ & $1.12\times^\dagger$ \\
\bottomrule
\end{tabular}}
\end{table}
\paragraph{More accurate formats on the same kernels.} Switching formats changes only the packer. On Qwen3-1.7B perplexity on the kernels is 10.387 (16-wide) and 10.271 (8-wide) against references 10.394 and 10.268, and prefill speed is unchanged: in the same sessions the 8-wide and 16-wide formats are $1.41$--$1.72\times$ MLX at 512--8192 tokens, SC-MXFP4 $1.43$--$1.72\times$ (Appendix~\ref{app:fmtsys}).

\paragraph{Tokens per step.} Speculative decoding and small batches run steps of a few tokens. On Qwen3.5-35B-A3B (Table~\ref{tab:tokens}), steps of 2--32 tokens run our prefill path, whose fixed cost (about 18\,ms) makes them 0.43--0.96$\times$ MLX's; at 48--96 tokens we are 1.08--1.60$\times$ faster, largely because MLX's own time rises steeply there, at 128--256 tokens the two are tied (0.95--1.07$\times$), and at 384--512 tokens we are 1.08--1.19$\times$ faster. A multi-row GEMV would help below 48 tokens; we have not built it.

\section{A 125B mixture of experts: Qwen3.8-Flash-Next}
\label{sec:q38}

Qwen3.8-Flash-Next, released in August 2026~\citep{qwen38}, has 125B language-model parameters: 48 layers of Gated DeltaNet and
sparse attention (3:1), each with 512 routed experts of $2560\times640$ (top-10) and a shared expert, four gated
residual streams, and a 51B-parameter n-gram embedding table that is only ever indexed. We start from the official
FP8 checkpoint (173\,GB; routed experts and the n-gram table in FP8, the rest BF16) and run it with the model code of
mlx-lm pull request~1788~\citep{mlxlmpr1788}. In BF16 the model does not fit in 128\,GB, so perplexity is evaluated
layer by layer: the hidden states of all 145 WikiText-2 windows are kept for every method, each layer is read once
from the checkpoint, quantized once per format and applied to every method's states. W4A8 covers every DeltaNet and
attention projection and every routed and shared expert (98\% of the parameters outside the n-gram table); the
router, gates, mixers and n-gram table stay as in the checkpoint. All formats use round-to-nearest without
calibration data.
\begin{table}[t]
\centering
\small
\caption{Qwen3.8-Flash-Next, WikiText-2 test perplexity (145 windows of 2048 tokens; FP8 checkpoint: 4.662). All
methods quantize the same matrices (every projection and expert); NVIDIA's NVFP4 release quantizes only routed experts~\citep{nvfp4q38}. bits/w: for this model's matrices (row scales add $32/K$; expert down projections have $K=640$). ``vs.\ FP8'': W4A8 column from unrounded perplexities; ``--'': not run (INT8 acts cost $\le 0.3\%$ elsewhere). Perplexities use the reference DeltaNet recurrence. Last column: execution on M5 ($^\ast$INT8 with 64-element group correction, oMLX~\citep{omlx}).}
\label{tab:q38}
\setlength{\tabcolsep}{4pt}
\resizebox{\columnwidth}{!}{%
\begin{tabular}{lrrrrl}
\toprule
Format & bits/w & W4A16 & W4A8 & vs.\ FP8 & on M5 \\
\midrule
\textbf{8-wide, $m\le10$ (ours)} & 4.53 & -- & \textbf{4.842} & $+3.9\%$ & \textbf{INT8, exact} \\
NVFP4, E4M3 scale search          & 4.50 & 4.904 & 4.917 & $+5.5\%$ & FP16 \\
NVFP4, amax/6 rule                & 4.50 & 4.918 & 4.928 & $+5.7\%$ & FP16 \\
16-wide, $m\le10$ (ours) & 4.27 & 4.970 & 4.969 & $+6.6\%$ & \textbf{INT8, exact} \\
Affine INT4, g64 (MLX 4-bit) & 4.50 & 4.995 & 4.996 & $+7.2\%$ & FP16 / INT8$^\ast$ \\
SC-MXFP4 (ours, std.\ MXFP4)   & 4.09 & 5.097 & 5.107 & $+9.6\%$ & \textbf{INT8, exact} \\
MXFP4, OCP rule (MLX)    & 4.25 & 5.199 & 5.212 & $+11.8\%$ & FP16 \\
\midrule
\multicolumn{6}{l}{\emph{Routed experts only (the scope of the public NVFP4 and FP8 releases)}} \\
\textbf{8-wide, $m\le10$ (ours)} & 4.53 & -- & \textbf{4.658} & $-0.1\%$ & \textbf{INT8, exact} \\
NVFP4, E4M3 scale search          & 4.50 & 4.656 & 4.661 & $-0.0\%$ & FP16 \\
16-wide, $m\le10$ (ours) & 4.28 & 4.668 & 4.664 & $+0.1\%$ & \textbf{INT8, exact} \\
SC-MXFP4 (ours, std.\ MXFP4)   & 4.09 & 4.666 & 4.671 & $+0.2\%$ & \textbf{INT8, exact} \\
NVFP4, amax/6 rule                & 4.50 & 4.674 & 4.678 & $+0.4\%$ & FP16 \\
Affine INT4, g64 (MLX 4-bit) & 4.50 & 4.754 & 4.761 & $+2.1\%$ & FP16 / INT8$^\ast$ \\
MXFP4, OCP rule (MLX)    & 4.25 & 4.621 & 4.619 & $-0.9\%$ & FP16 \\
\bottomrule
\end{tabular}}
\end{table}
\paragraph{Accuracy.} Table~\ref{tab:q38} reports perplexity on Qwen3.8-Flash-Next. INT8 activations cost $\le 0.3\%$ for every format; the 4-bit loss sits in the dense projections (2--3\% of weights), whereas quantizing only the 73{,}728 routed experts costs $\le 0.4\%$ ($\le 0.2\%$ for ours). There perplexity stops separating formats (OCP MXFP4 even scores below FP8), but KL divergence from FP8 separates them (32 windows, INT8 activations): $0.0557$ (16-wide), $0.0553$ (NVFP4 search), $0.0628$ (affine INT4), $0.0802$ (SC-MXFP4), and $0.0940$ (OCP MXFP4; Appendix~\ref{app:q38speed}). With all matrices quantized, 8-wide is most accurate (4.842 vs.\ 4.917 NVFP4 search, $-0.075$ $[-0.095,-0.056]$, and 4.996 affine INT4), and 16-wide (4.27 bits) beats both deployed Mac formats. Mixed dense projections help more than on 35B (Appendix~\ref{app:scope}): 16-wide experts reach 4.711 with 8-bit dense projections ($-0.131$) and 4.852 with 8-wide ones (tied, 0.25 fewer bits). Row compliance is 99.929\% across 95.4M rows; packed weights match fake quantization (0/298M differ), and kernel perplexity matches within 0.4\% on 8 windows. On 1319 GSM8K problems (5-shot, greedy), W4A8 solves 92.65$\pm$0.72\% (SC-MXFP4), 94.84$\pm$0.61\% (16-wide), and 95.30$\pm$0.58\% (8-wide) vs.\ 92.80$\pm$0.71\% (MLX on SC-MXFP4) and 90.98$\pm$0.79\% (MLX MXFP4; 8-wide vs.\ MLX MXFP4: $p<10^{-9}$; vs.\ SC-MXFP4: $p=10^{-4}$; INT8 acts: $p=0.87$).
\paragraph{Speed.} In Table~\ref{tab:q38speed} (Appendix~\ref{app:q38speed}), prefill runs $1.27$--$1.35\times$ faster than MLX MXFP4 at 512--2048 tokens and $1.18$--$1.22\times$ at 8192 ($1.33$--$1.65\times$ on isolated MoE blocks; $\sim0.5$\,s is shared outside W4A8). Streaming-bound batch-1 decode matches MLX ($1.03\times$ SC-MXFP4 using MLX dense GEMVs; $1.01$ and $0.98\times$ for 16- and 8-wide). Vs.\ oMLX's per-group kernels~\citep{omlx}, dense projections run $0.89$--$1.24\times$ and experts $0.76$--$0.94\times$ (its down projection stays W4A16).

\section{Related work}\label{sec:related}

\paragraph{W4A8 and integer scales.} Outlier-aware 8/4-bit methods establish W4A16 compression~\citep{llmint8,awq,qlora,marlin}. For W4A8, VS-Quant~\citep{vsquant} pairs float row and integer block scales (like llama.cpp k-quants and IQ4\_KS~\citep{llamacpp,iq4ks}); QQQ~\citep{qqq} and Integer Scale~\citep{integerscale} apply integer scales to partial sums, whereas QServe~\citep{qserve} and DGQ~\citep{dgq} bound INT4$\times$scale inside INT8 for one GEMM. We use E2M1 and $m\le10$ inside that budget, cutting 1.7B output error from $12.38$ to $7.36\times10^{-3}$ (Table~\ref{tab:formats}).

\paragraph{FP4 on integer hardware.} Existing kernels apply E8M0 scales to partial sums~\citep{llamacppfp4,mlspecqd,vllmw4a8mxfp4,blockdialect}; ZeroQuant-FP enforces power-of-two shifts~\citep{zeroquantfp}, shared microexponents window sub-block shifts~\citep{sharedmicroexp}, and AdaMX exploits narrow channel ranges in custom silicon~\citep{adamx}. We give the exact row-level folding condition and quantizer.

\paragraph{FP4 formats.} Beyond OCP MXFP4~\citep{ocpmx} and NVFP4~\citep{nvfp4}, MR-GPTQ~\citep{mrgptq}, Four Over Six~\citep{fourover6}, and scale searches~\citep{scalesearch,scalesweep,soscale,osfp4} refine block scales, and \citet{chhugani2026mxfp4} rescale 128-element partial sums; alternate formats~\citep{m2xfp,nxfp,hifloat4,mixfp4,if4} require new hardware rather than INT8 units. Because dense projections dominate MoE loss (\S\ref{sec:acc}, \S\ref{sec:q38}), expert-balanced, low-rank, mixed-precision, and KV-compression MoE methods~\citep{milo,moequant,mxmoe,rotaryquant} compose with us, plus SmoothQuant~\citep{smoothquant}, Sinkhorn~\citep{sinq}, and rotations~\citep{quarot,spinquant}.

\paragraph{Apple Silicon.} BaseRT~\citep{basertm5} runs on M5 without operand details. Cider~\citep{cider} runs per-channel INT4$\times$INT8, while oMLX~\citep{omlx,omlxmx} and scratchy~\citep{scratchy} correct partial sums per group; folding scales yields exact cross-vendor GEMMs and stronger FP4 formats (Appendix~\ref{app:related}).

\section{Limitations and conclusion}
\label{sec:concl}

End-to-end timing covers one M5 Max and one NVIDIA L4 (A100 is layer-level). GPTQ, bypass, SmoothQuant and per-class formats are simulated. With 8-wide blocks, decode is 2--10\% slower than MLX; steps of 2--32 tokens are slower, and 128--256 tokens are tied. Folding scales buys exactness and stronger formats rather than speed over concurrent per-group kernels (Appendix~\ref{app:limits}).

On INT8$\times$INT8 units, 4-bit formats need not trade away accurate block scales: integer multipliers keep each row an exact INT8 vector with one scale, and $m\le10$ for E2M1 beats calibration-free NVFP4 search on all eight models (plus calibrated and W4A16 NVFP4 across three dense models under folded SmoothQuant) at the INT8 rate.

\paragraph{Reproducibility.} The code repository, \url{https://github.com/shinyoungkang1/mxfp4-int8-m5}, contains
every kernel, harness and evaluation script, from which all reported numbers come; all models and datasets are public.

\section*{Impact Statement}\label{sec:impact}

This work lowers the cost of running large language models on personal hardware, which can improve privacy and reduce
dependence on hosted services. It does not change what the models can do, and local deployment also removes the
safeguards a hosted service can apply; the formats inherit the risks and licenses of the models they compress. End-to-end speed
is measured on one M5 Max and one L4 and may not transfer without measurement.

\section*{AI Disclosure Statement}\label{sec:aidisclosure}

The authors conceived the research problem, hypotheses, format family, mathematical analysis, and experimental design. Under the authors' direction, generative AI coding assistants (Anthropic's Claude and Google's Gemini) assisted in implementing the kernels, benchmark harnesses, and quantization and evaluation scripts, as well as in editing and polishing the manuscript. The authors reviewed, tested, and verified all code, proofs, experimental results, references, and text, and take full responsibility for the content of this paper.

\bibliography{refs}
\bibliographystyle{iclr2027_conference}

\appendix

\section{Proofs}
\label{app:proofs}

\paragraph{Proposition~\ref{prop:ladder}.} Since $0<a\le A$, $0<a/(\beta C)\le A/(\beta C)=M$, so $m\in\{1,\ldots,M\}$. Then
$s=\beta m\ge a/C$, so no element of the block exceeds the codebook's top, and $s<\beta(a/(\beta C)+1)=a/C+\beta=(a/C)(1+A/(Ma))$,
using $\beta C=A/M$. Round-to-nearest errs by at most half the local gap times $s$, hence by at most $gs/2$. The quantizer of \S\ref{sec:family} picks the multiplier with the least squared error and searches $\beta$ per row, so its row's total squared error is at most that of this choice. For $\{1,2,4,8\}$, a block whose peak is just above $2^iC$ times the base must use $2^{i+1}$, with scale just below $2a/C$.

\paragraph{Proposition~\ref{prop:embed}.} $2v_{n,k}$ is an integer and $e_{n,b(k)} - E_n$ is a non-negative
integer, so $q_{n,k}$ is an integer, and $2^{E_n-1}q_{n,k} = 2^{e_{n,b(k)}}v_{n,k}$ term by term. Since $|2v| \le 12$
and $e_{n,b(k)} - E_n \le s_n$, $|q_{n,k}| \le 12\cdot 2^{s_n} \le 96$ for $s_n \le 3$. If $s_n \ge 4$ and a block at
exponent $E_n + s_n$ contains $v = \pm4$ or $\pm6$, then $|q| \ge 8\cdot 2^{s_n} \ge 128$, outside the symmetric INT8 range. Under the OCP rule a
block's maximum lies in $[4\cdot2^{e_b}, 8\cdot2^{e_b})$ and rounds to $\pm4$ or $\pm6$.

\paragraph{Proposition~\ref{prop:cost}.} Let block $b$ have exponent $e_b \le E^* - 4$ before the constraint and
$L = E^* - 3$ after, and let $e_O = \lfloor\log_2\max_k|w_k|\rfloor - 2$ be its OCP exponent, so that
$|w| < 8\cdot 2^{e_O}$. Exponent search picks $e_b \in \{e_O-1, e_O, e_O+1\}$, so $e_O \le L-1$ except when
$e_b = L-1$ and $e_O = L$. If $e_O \le L-1$, every value satisfies $|w| < 8\cdot 2^{L-1} = 4\cdot 2^{L}$; for $|w|$ in
$[0, 4\cdot2^L]$ (the grid is symmetric) the scaled E2M1 grid $\{0, \tfrac12, 1, \tfrac32, 2, 3, 4\}\cdot 2^L$ has spacing at most $2^L$, so
round-to-nearest at exponent $L$ errs by at most $2^{L-1} = 2^{E^*-4}$. In the remaining case the search preferred
exponent $L-1$ to $L$ although the block maximum lies in $[4\cdot2^L, 8\cdot2^L)$. The two grids coincide on
$[2^L, 3\cdot2^L]$; below $2^L$ the finer grid of $L-1$ lowers an element's squared error by at most
$(2^{L-2})^2 = 2^{2L}/16$, and above $3\cdot2^L$ it clips to $3\cdot2^L$, which never lowers the squared error: on
$(3, \tfrac72]\cdot2^L$ both grids round to $3\cdot2^L$; on $(\tfrac72, 5]\cdot2^L$, where exponent $L$ rounds to
$4\cdot2^L$, clipping raises it by $(2|w| - 7\cdot2^L)\,2^L$; on $(5, 8)\cdot2^L$, where $L$ rounds or saturates to
$6\cdot2^L$, by $3\cdot2^L(2|w| - 9\cdot2^L) > 3\cdot2^{2L}$.
At most 31 elements can gain, so $L-1$ wins only if every element above $4\cdot2^L$ has
$|w| \le \tfrac{143}{32}\cdot2^L$; at exponent $L$ such an element rounds to $4\cdot2^L$ with error at most
$\tfrac{15}{32}\cdot2^L < 2^{L-1}$, and the other elements are below $4\cdot2^L$, covered above. In a block at
exponent $E^*$, the grid spacing between 4 and 6 is $2\cdot 2^{E^*}$, so round-to-nearest can err by up to $2^{E^*}$
between grid points (values above $6\cdot2^{E^*}$ saturate and can err more).

\section{The format space}
\label{app:space}

\texttt{sim/format\_space.py} scores every combination of eleven codebook shapes (doubled E2M1, uniform INT4, five hand-picked shapes and four Lloyd shapes rounded to integers), block widths 8, 16, 32 and 64 and three multiplier sets using the whole INT8 budget (all integers, eight, or four, roughly geometric), and $M=1,\ldots,10$ for E2M1, 134 formats in all, on Qwen3-1.7B, Qwen3-4B and SmolLM3 (every third layer), with an oracle for each shape and width that gives each block a free float scale (32 candidates over two octaves) and is not INT8-constrained. \texttt{results/16\_space\_*.csv} hold all of them. Weight error relative to that oracle at 16-wide blocks falls as $1.79, 1.61, 1.46, 1.32, 1.17, 1.12, 1.08, 1.05, 1.02, 1.00$ for $M=1,\ldots,10$ (Qwen3-1.7B; the other two models agree within 0.04), against the scale-excess factor of \S\ref{sec:space}, $3.31, 2.16, 1.69, 1.49, 1.39, 1.31, 1.27, 1.23, 1.20, 1.18$; the factor is computed from block peaks alone and is not itself a bound on squared error, which stays below it at every $M$.

\begin{table}[h]
  \centering\small
  \caption{Integer-exact formats (Eq.~\ref{eq:family}) vs.\ standard 4-bit formats on Qwen3-1.7B: layer-output error
  $\mathrm{tr}(\Delta W H \Delta W^\top)/\mathrm{tr}(W H W^\top)$ with $H$ from WikiText-2 \emph{train} activations,
  every third layer, round-to-nearest. Bits include block codes and one FP32 scale per row ($K=4096$).
  ``+ rotation'': randomized Hadamard transform of each 32-wide input block. Scale search (a simple form of~\citep{scalesearch,scalesweep}): each NVFP4 block's E4M3
  scale and its two neighbors on either side; exponent search: the OCP exponent and its two neighbors (least squared
  weight error).}
  \label{tab:formats}
  \resizebox{\columnwidth}{!}{\begin{tabular}{lrrc}
\toprule
Format & Bits & Output error ($\times10^3$) & INT8-exact \\
\midrule
NVFP4, scale search & 4.50 & 5.70 & no \\
E2M1, 8-blocks, $m\in\{1..10\}$ & 4.51 & 5.99 & yes \\
NVFP4 (amax/6 rule) & 4.50 & 6.02 & no \\
E2M1, 16-blocks, $m\in\{1..10\}$ & 4.26 & 7.36 & yes \\
E2M1, 16-blocks, $m\in\{1..8\}$ & 4.19 & 7.97 & yes \\
E2M1, 32-blocks, $m\in\{1..8\}$ & 4.10 & 8.88 & yes \\
E2M1, 32-blocks, $m\in\{1..10\}$ & 4.13 & 8.99 & yes \\
MXFP4 (OCP rule) & 4.25 & 9.86 & no \\
INT4, 16-blocks, $m\in\{1..16\}$ + rotation & 4.26 & 10.52 & yes \\
MXFP4, exponent search & 4.25 & 10.59 & no \\
SC-MXFP4 (standard MXFP4 file) & 4.07 & 10.61 & yes \\
E2M1, 16-blocks, $m\in\{1,2,4,8\}$ & 4.13 & 11.68 & yes \\
INT4, 16-blocks, $m\in\{1..16\}$ & 4.26 & 12.38 & yes \\
E2M1, 16-blocks, $m\in\{1..8\}$ + rotation & 4.19 & 12.42 & yes \\
INT4, 16-blocks, $m\in\{1..8\}$ & 4.19 & 13.03 & yes \\
E2M1, 32-blocks, $m\in\{1..10\}$ + rotation & 4.13 & 14.01 & yes \\
E2M1, 32-blocks, $m\in\{1,2,4,8\}$ (float row scale) & 4.07 & 14.07 & yes \\
NVFP4 + rotation & 4.50 & 14.17 & no \\
INT4 affine g64 (MLX) & 4.50 & 14.27 & no \\
INT4, 32-blocks, $m\in\{1..16\}$ & 4.13 & 15.12 & yes \\
SC-MXFP4 (standard MXFP4 file) + rotation & 4.07 & 19.45 & yes \\
\bottomrule
\end{tabular}
}
\end{table}

\section{Confidence intervals}
\label{app:ci}

\begin{table}[h]
\centering\small
\caption{Perplexity differences, WikiText-2, INT8 activations, fake quantization: point estimate and 95\% paired bootstrap interval (10000 resamples of the same 140--163 windows for both methods; bold: excludes zero). Windows are neighboring text from one test set, so this is an interval over that set, not over text in general. \texttt{results/17\_bootstrap.csv}, \texttt{results/54\_bootstrap\_olmoe\_b8.csv}, and \texttt{results/57\_mistral7b\_bootstrap.csv}.}
\label{tab:ci}
\setlength{\tabcolsep}{3pt}
\resizebox{\textwidth}{!}{%
\begin{tabular}{lrrrrrr}
\toprule
Difference & Qwen3-0.6B & Qwen3-1.7B & Qwen3-4B & SmolLM3 & Mistral-7B & OLMoE \\
\midrule
16-wide $-$ SC-MXFP4 & \textbf{-0.959 [-1.044, -0.880]} & \textbf{-0.596 [-0.644, -0.549]} & \textbf{-0.322 [-0.351, -0.293]} & \textbf{-0.275 [-0.296, -0.254]} & \textbf{-0.074 [-0.083, -0.066]} & \textbf{-0.262 [-0.297, -0.231]} \\
16-wide $-$ NVFP4 & \textbf{-0.318 [-0.359, -0.276]} & \textbf{-0.114 [-0.138, -0.089]} & \textbf{-0.049 [-0.065, -0.033]} & -0.010 [-0.023, +0.003] & +0.006 [-0.001, +0.012] & \textbf{-0.072 [-0.083, -0.060]} \\
16-wide $-$ NVFP4, search & \textbf{-0.034 [-0.066, -0.004]} & \textbf{+0.058 [+0.034, +0.084]} & -0.015 [-0.032, +0.003] & \textbf{+0.045 [+0.033, +0.058]} & \textbf{+0.026 [+0.020, +0.032]} & \textbf{+0.014 [+0.002, +0.025]} \\
8-wide $-$ NVFP4, search & \textbf{-0.437 [-0.485, -0.392]} & \textbf{-0.069 [-0.094, -0.043]} & \textbf{-0.070 [-0.089, -0.052]} & \textbf{-0.048 [-0.060, -0.036]} & \textbf{-0.013 [-0.019, -0.006]} & \textbf{-0.070 [-0.083, -0.057]} \\
8-wide $-$ NVFP4 & \textbf{-0.720 [-0.781, -0.662]} & \textbf{-0.240 [-0.269, -0.212]} & \textbf{-0.104 [-0.124, -0.086]} & \textbf{-0.103 [-0.118, -0.090]} & \textbf{-0.033 [-0.040, -0.026]} & \textbf{-0.156 [-0.170, -0.141]} \\
8-wide $\{..,10\},m\le12$ $-$ 8-wide & \textbf{-0.181 [-0.215, -0.148]} & \textbf{-0.025 [-0.049, -0.001]} & +0.017 [-0.000, +0.034] & \textbf{-0.024 [-0.035, -0.013]} & -- & -- \\
8-wide $\{..,10\},m\le12$ $-$ NVFP4, search & \textbf{-0.617 [-0.670, -0.569]} & \textbf{-0.093 [-0.121, -0.066]} & \textbf{-0.054 [-0.074, -0.033]} & \textbf{-0.072 [-0.086, -0.059]} & -- & -- \\
8-wide $-$ 16-wide & \textbf{-0.402 [-0.444, -0.362]} & \textbf{-0.127 [-0.151, -0.103]} & \textbf{-0.056 [-0.072, -0.040]} & \textbf{-0.094 [-0.107, -0.080]} & \textbf{-0.038 [-0.045, -0.032]} & \textbf{-0.084 [-0.097, -0.071]} \\
16-wide $\{..,10\},m\le12$ $-$ 16-wide & \textbf{-0.161 [-0.203, -0.122]} & \textbf{+0.376 [+0.341, +0.410]} & \textbf{+0.038 [+0.018, +0.058]} & -0.013 [-0.027, +0.000] & -- & -- \\
16-wide $\{..,14\},m\le9$ $-$ 16-wide & -0.022 [-0.061, +0.015] & +0.026 [-0.004, +0.054] & \textbf{+0.070 [+0.051, +0.090]} & \textbf{-0.026 [-0.038, -0.014]} & -- & -- \\
32-wide, $m\le8$ $-$ NVFP4 & +0.027 [-0.023, +0.077] & \textbf{+0.348 [+0.308, +0.391]} & \textbf{+0.098 [+0.079, +0.119]} & \textbf{+0.059 [+0.045, +0.072]} & \textbf{+0.020 [+0.013, +0.026]} & -- \\
\bottomrule
\end{tabular}}
\end{table}

\section{Hardware characterization}
\label{app:micro}

\paragraph{A second kind of matrix unit (Arm SME).} On the M5 CPU's SME unit (512-bit vectors, one thread), INT8 SMOPA on real Qwen3-1.7B operands equals the exact integer product on all six layer/format pairs we ran, and takes $0.27$--$0.30\times$ the time of widening FP16 and of FP32 FMOPA on the same matrices ($3.3$--$3.6\times$ faster, $M=256$ synthetic activation rows, $K$ and $N$ up to 6144); its INT8 outer product computes four products per output element and instruction, FP16 two. This is another unit on the same chip, not another vendor; we did not run AMD or Intel hardware.

\paragraph{NVIDIA INT8 Tensor Cores.} Table~\ref{tab:nvidia} reports single-layer benchmarks (\texttt{bench/nvidia\_int8\_bench.py}) on NVIDIA A100-SXM4-40GB (\texttt{sm\_80}, PyTorch~2.14.1+cu130, Triton~3.8.0) and NVIDIA L4 (\texttt{sm\_89}), with the L2 cache flushed before every timed call (median of 15 interleaved repetitions). Across all ten shapes (the six prefill shapes in Table~\ref{tab:nvidia}, small-batch $M\in\{32,128\}$ at $4096\times4096$, and MoE dense shapes $2048\times6144\times2048$ and $2048\times2560\times6144$) and all three formats (\texttt{SC-MXFP4}, 16-wide, 8-wide), Triton unpacking has 0 mismatches, INT32 accumulation has 0 errors against \texttt{int64}, and our autotuned Triton fused-epilogue INT8 \texttt{tl.dot} kernel matches \texttt{torch.\_int\_mm} plus separate FP16 rescale with 0 FP16 mismatches; Triton K-loop \texttt{W4A16 fused} has $3.3$--$4.3\times10^{-4}$ relative $L_2$ error vs.\ FP16. Latency across the three integer-exact formats agrees within 1--4\% across Table~\ref{tab:nvidia} (at $M=512$ on A100, 8-wide and 16-wide at $0.121$--$0.123$\,ms are 8--9\% faster than SC-MXFP4 at $0.133$\,ms).

\begin{table}[h]
\centering\small
\caption{Linear-layer latency (ms, median of 15 interleaved repetitions, L2 flushed) on NVIDIA A100 and L4 using the same SC-MXFP4 weights and Table~\ref{tab:layer} shapes (\texttt{results/nvidia/nvidia\_int8\_\{a100\_sxm4\_40gb,l4\}.csv}). Marlin W4A16~\citep{marlin} uses \texttt{marlin.Layer(g128)}. \texttt{W4A8} reports the faster of two-pass (\texttt{torch.\_int\_mm} + Triton rescale) and Triton fused-epilogue INT8 GEMM ($^\star$); \texttt{W8A8}$^\dagger$ assumes pre-unpacked INT8 weights. Speedups are medians of per-repetition ratios. The 16-wide and 8-wide formats run on the same kernel at nearly identical latency.}
\label{tab:nvidia}
\setlength{\tabcolsep}{5pt}\begin{tabular}{lrrrrrrrr}
\toprule
 & cuBLAS & Marlin & W4A16 & W4A16 & \textbf{W4A8} & & \multicolumn{2}{c}{W4A8 vs.} \\
\cmidrule(lr){8-9}
$M\times N\times K$ & FP16 & W4A16 & fused & 2-pass & \textbf{(ours)} & W8A8$^\dagger$ & FP16 & Marlin \\
\midrule
\multicolumn{9}{l}{\emph{NVIDIA A100-SXM4-40GB (\texttt{sm\_80}, 40\,GB; \texttt{torch 2.14+cu130})}} \\
512$\times$4096$\times$4096 & \textbf{0.092} & 0.100 & 0.295 & 0.129 & 0.133 & 0.102 & 0.70$\times$ & 0.76$\times$ \\
1024$\times$4096$\times$4096 & \textbf{0.157} & 0.182 & 0.480 & 0.195 & 0.174 & 0.140 & 0.90$\times$ & 1.05$\times$ \\
2048$\times$4096$\times$4096 & \textbf{0.261} & 0.358 & 0.777 & 0.300 & 0.279 & 0.244 & 0.94$\times$ & 1.29$\times$ \\
4096$\times$4096$\times$4096 & 0.527 & 0.709 & 1.494 & 0.566 & \textbf{0.490} & 0.461 & 1.08$\times$ & 1.44$\times$ \\
2048$\times$14336$\times$4096 & 0.969 & 1.155 & 2.533 & 1.096 & \textbf{0.833} & 0.739 & 1.17$\times$ & 1.39$\times$ \\
2048$\times$4096$\times$14336 & 0.889 & 1.151 & 2.590 & 1.018 & \textbf{0.834} & 0.736 & 1.07$\times$ & 1.38$\times$ \\
\midrule
\multicolumn{9}{l}{\emph{NVIDIA L4 (\texttt{sm\_89}, 24\,GB; $^\star$Triton fused-epilogue INT8 GEMM when faster)}} \\
512$\times$4096$\times$4096 & 0.354 & 0.310 & 0.507 & 0.465 & \textbf{0.309} & 0.200$^\star$ & 1.15$\times$ & 1.00$\times$ \\
1024$\times$4096$\times$4096 & 0.563 & 0.548 & 0.908 & 0.702 & \textbf{0.472$^\star$} & 0.368$^\star$ & 1.19$\times$ & 1.16$\times$ \\
2048$\times$4096$\times$4096 & 1.012 & 1.093 & 1.881 & 1.163 & \textbf{0.880} & 0.786 & 1.15$\times$ & 1.25$\times$ \\
4096$\times$4096$\times$4096 & 1.863 & 2.195 & 3.800 & 2.035 & \textbf{1.686$^\star$} & 1.584$^\star$ & 1.11$\times$ & 1.30$\times$ \\
2048$\times$14336$\times$4096 & 3.282 & 4.192 & 6.651 & 3.924 & \textbf{2.952$^\star$} & 2.465$^\star$ & 1.11$\times$ & 1.42$\times$ \\
2048$\times$4096$\times$14336 & 3.472 & 4.221 & 6.694 & 4.024 & \textbf{2.861} & 2.376 & 1.22$\times$ & 1.48$\times$ \\
\bottomrule
\end{tabular}

\end{table}

On A100 (\texttt{sm\_80}, 1.55\,TB/s HBM, cuBLASLt INT8 reaching up to 421 TOPS isolated), cuBLASLt two-pass outperforms Triton fused-epilogue at $M\ge512$, and \texttt{W4A8} beats \texttt{W4A16 fused} ($2.20$--$3.11\times$), \texttt{W4A16 2-pass} ($1.08$--$1.32\times$ at $M\ge1024$), \texttt{Marlin W4A16} ($1.05$--$1.44\times$ at $M\ge1024$), and \texttt{cuBLAS FP16} ($1.07$--$1.17\times$ on the largest three shapes). On L4 (\texttt{sm\_89}, 300\,GB/s HBM), fusing the row-scale epilogue into Triton's INT8 \texttt{tl.dot} eliminates the $0.07$--$0.82$\,ms INT32 global write/read and wins on 4 of 6 shapes, so \texttt{W4A8} beats all four baselines across Table~\ref{tab:nvidia} ($1.11$--$1.22\times$ vs.\ FP16, $1.00$--$1.48\times$ vs.\ Marlin, $1.20$--$1.51\times$ vs.\ \texttt{W4A16 2-pass}, $1.65$--$2.34\times$ vs.\ \texttt{W4A16 fused}). PyTorch's decode-oriented \texttt{\_weight\_int4pack\_mm} stays at 18 TFLOPS (A100) and 7--8.4 TFLOPS (L4) at $M\ge512$ ($6.6$--$16.0\times$ slower than W4A8). On MoE dense shapes ($2048\times6144\times2048$, $2048\times2560\times6144$), \texttt{W4A8} beats Marlin by $1.02$--$1.22\times$ on A100 and $1.21$--$1.39\times$ on L4; at small batch ($M=32,128$), weight bandwidth dominates and Marlin leads ($0.031$/$0.042$\,ms on A100, $0.060$/$0.112$\,ms on L4), matching the decode bound on M5.

\paragraph{End-to-end NVIDIA L4 prefill and kernel accuracy.} Table~\ref{tab:nvidia_e2e} wraps the same verified Triton unpack + INT8 \texttt{tl.dot} / \texttt{torch.\_int\_mm} \texttt{W4A8Linear} modules (\texttt{bench/nvidia\_e2e\_bench.py}) end-to-end into \texttt{Qwen3-4B-Base} and \texttt{Mistral-7B-v0.3} on NVIDIA L4 (\texttt{sm\_89}, 24\,GB; A100 remains layer-level in Table~\ref{tab:nvidia}). Checking layer~0 \texttt{q\_proj} on both models and formats finds 0 unpack mismatches, 0 INT32 GEMM errors vs.\ \texttt{int64}, and 0 fused-vs-unfused FP16 mismatches, with quantized weight-MSE parity $\le 8.2\times10^{-10}$ against NumPy reference quantization. End-to-end WikiText-2 kernel perplexity matches fake quantization within $0.011$ (Qwen3-4B: 8.742 vs.\ 8.745 for SC-MXFP4, 8.357 vs.\ 8.368 for 8-wide; Mistral-7B: 5.592 vs.\ 5.592 and 5.479 vs.\ 5.480). On Mistral-7B, W4A8 cuts peak VRAM from $14.0$\,GiB (FP16) to $4.7$--$5.1$\,GiB ($2.7$--$3.0\times$) and beats both FP16 ($1.18$--$1.35\times$) and Marlin W4A16 ($1.22$--$1.68\times$) across all prompt lengths $L=512$--$4096$; on Qwen3-4B it cuts peak VRAM from $7.8$\,GiB to $3.0$--$3.2$\,GiB ($2.4$--$2.6\times$) and leads at $L\ge1024$ ($1.18$--$1.37\times$ vs.\ FP16, $1.29$--$1.46\times$ vs.\ Marlin), trailing at $L=512$ ($0.90\times$ / $0.85$--$0.86\times$) due to fixed activation-quantization and unpack overhead.

\begin{table}[h]
\centering\small
\caption{End-to-end NVIDIA L4 (\texttt{sm\_89}, PyTorch~2.14.1+cu130, Triton~3.8.0) prefill TTFT (ms, median of 15 repetitions, L2 flushed), peak VRAM at $L=4096$, and full WikiText-2 kernel perplexity vs.\ fake-quant simulation (\texttt{results/nvidia/e2e\_nvidia\_l4.csv}). Marlin uses \texttt{marlin.Layer(g128)}.}
\label{tab:nvidia_e2e}
\setlength{\tabcolsep}{4pt}\resizebox{\linewidth}{!}{\begin{tabular}{llrrrrrrrr}
\toprule
 & & Wt / Peak & PPL & \multicolumn{4}{c}{Prefill TTFT (ms)} & \multicolumn{2}{c}{W4A8 vs.} \\
\cmidrule(lr){5-8}\cmidrule(lr){9-10}
Model & System & (GiB) & (sim) & 512 & 1024 & 2048 & 4096 & FP16 & Marlin \\
\midrule
Qwen3-4B & FP16 & 7.49 / 7.80 & 7.894 (7.898) & 100.4 & 226.3 & 399.8 & 927.1 & -- & -- \\
         & Marlin W4A16 & 2.47 / 2.78 & -- & \textbf{95.1} & 216.5 & 489.3 & 1130.6 & 0.82--1.06$\times$ & 1.00$\times$ \\
         & W4A8, SC-MXFP4 & 2.45 / 3.03 & 8.742 (8.745) & 111.6 & 168.1 & 337.6 & \textbf{778.5} & 0.90--1.35$\times$ & 0.85--1.45$\times$ \\
         & \textbf{W4A8, 8-wide} & 2.63 / 3.22 & 8.357 (8.368) & 111.0 & \textbf{165.7} & \textbf{335.4} & 782.7 & 0.90--1.37$\times$ & 0.86--1.46$\times$ \\
\midrule
Mistral-7B & FP16 & 13.50 / 13.96 & 5.318 (5.318) & 179.9 & 330.4 & 684.4 & 1422.3 & -- & -- \\
           & Marlin W4A16 & 3.85 / 4.32 & -- & 186.8 & 396.5 & 854.2 & 1801.3 & 0.79--0.96$\times$ & 1.00$\times$ \\
           & W4A8, SC-MXFP4 & 3.81 / 4.73 & 5.592 (5.592) & \textbf{150.6} & \textbf{263.1} & 511.7 & \textbf{1080.9} & 1.20--1.34$\times$ & 1.24--1.67$\times$ \\
           & \textbf{W4A8, 8-wide} & 4.16 / 5.09 & 5.479 (5.480) & 153.1 & 264.2 & \textbf{508.5} & 1091.4 & 1.18--1.35$\times$ & 1.22--1.68$\times$ \\
\bottomrule
\end{tabular}
}
\end{table}

\begin{figure}[t]
  \centering
  \includegraphics[width=0.8\textwidth]{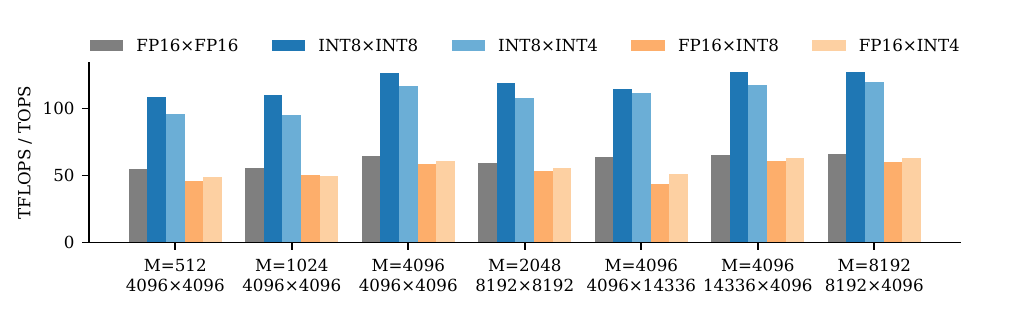}
  \caption{\texttt{matmul2d} throughput on the M5 Max Neural Accelerators by operand type (left $\times$ right
  operand) for $M\times N\times K$ shapes. INT8$\times$INT8 reaches $1.8$--$2.0\times$ FP16; every mixed
  float$\times$integer combination is slower than FP16.}
  \label{fig:micro}
\end{figure}

\begin{table}[h]
\centering\small
\caption{\texttt{matmul2d} throughput (TFLOPS or TOPS, median of 15 interleaved repetitions) on the M5 Max, AC
power, High Power mode; speedup vs.\ FP16 in parentheses (median of per-repetition ratios). One \texttt{matmul2d}
call per $128\times64$ output tile, 4 simdgroups, grouped tile order, weights $N\times K$. All outputs bit-exact.}
\setlength{\tabcolsep}{3pt}
\begin{tabular}{lrrrrr}
\toprule
$M\times N\times K$ & FP16 & INT8$\times$INT8 & INT8$\times$INT4 & FP16$\times$INT8 & FP16$\times$INT4 \\
\midrule
$512\times4096\times4096$ & 54.9 & 108.7 (1.97) & 95.5 (1.74) & 45.9 (0.84) & 48.9 (0.89) \\
$1024\times4096\times4096$ & 55.6 & 110.2 (1.98) & 94.9 (1.70) & 50.3 (0.90) & 49.5 (0.89) \\
$4096\times4096\times4096$ & 64.4 & 126.7 (1.97) & 116.3 (1.81) & 58.5 (0.91) & 60.8 (0.94) \\
$2048\times8192\times8192$ & 59.6 & 118.9 (1.98) & 108.1 (1.81) & 53.2 (0.90) & 55.8 (0.93) \\
$4096\times14336\times4096$ & 65.4 & 127.4 (1.95) & 117.8 (1.80) & 60.5 (0.92) & 62.8 (0.96) \\
$4096\times4096\times14336$ & 63.9 & 114.2 (1.79) & 111.5 (1.74) & 43.5 (0.68) & 50.7 (0.79) \\
$8192\times8192\times4096$ & 65.6 & 127.3 (1.94) & 119.3 (1.82) & 60.1 (0.92) & 62.8 (0.96) \\
\bottomrule
\end{tabular}
\end{table}

Storing weights $K\times N$ instead of $N\times K$ (reduction dimension contiguous) is about $2\times$ slower for every
operand type. FP16 output is no faster than FP32 output under sustained load. On battery with Low Power Mode,
sustained throughput drops to about a third (FP16 $\approx 23$ TFLOPS) with unchanged INT8/FP16 ratios
($1.85$--$2.04\times$), so the ratio is not a power-state artifact.

\paragraph{macOS 27.2 FP4, FP8, and MXFP4 operands.} Table~\ref{tab:macos27fp} repeats the same \texttt{matmul2d} microbenchmark on macOS~27.2 including the newly exposed FP8, FP4, and block-scaled MXFP4 (one E8M0 scale per 32 weights) operand types (\texttt{results/53\_microbench\_macos27\_fp.csv}, AC power, 15 interleaved repetitions, \texttt{max\_abs\_err} $= 0.0$ across all tested integer-valued inputs). E4M3 and E5M2 run at the FP16 rate ($0.97$--$1.00\times$ on the six clean shapes; \texttt{FP16}$\times$\texttt{FP8}: $0.96$--$1.00\times$), FP4 runs below FP16 ($0.78$--$0.85\times$), and attaching per-32 MXFP4 scales drops throughput to $0.41$--$0.46\times$ (\texttt{FP16}$\times$\texttt{MXFP4}) and $0.27$--$0.31\times$ (\texttt{MXFP4}$\times$\texttt{MXFP4}), while \texttt{INT8}$\times$\texttt{INT8} maintains $1.91$--$1.97\times$. Thus even on macOS~27.2, M5's $2\times$ speedup requires integer operands for both activations and weights.

\begin{table}[h]
\centering\small
\caption{\texttt{matmul2d} speedup vs.\ FP16$\times$FP16 (TFLOPS in column 2) on M5 Max under macOS~27.2 (AC power, High Power mode, median of 15 interleaved repetitions; \texttt{kernels/bench\_fp.swift}). \texttt{FP8} is E4M3 (E5M2 matches at $0.97$--$1.00\times$). On $4096\times4096\times14336$, the FP16 baseline dipped to 41.8\,TFLOPS, lifting ratios to $2.23\times$ (\texttt{INT8}$\times$\texttt{INT8}), $1.23\times$ (\texttt{FP8}), $1.01\times$ (\texttt{FP4}), $0.54\times$ (\texttt{FP16}$\times$\texttt{MXFP4}), and $0.34\times$ (\texttt{MXFP4}); \texttt{INT8}$\times$\texttt{INT8} still leads \texttt{FP4} by $2.2\times$ and \texttt{MXFP4} by $4.2$--$6.8\times$.}
\label{tab:macos27fp}
\setlength{\tabcolsep}{4pt}
\begin{tabular}{lrrrrrrr}
\toprule
$M\times N\times K$ & FP16 (TF) & INT8$\times$INT8 & FP8$\times$FP8 & FP4$\times$FP4 & FP16$\times$FP4 & FP16$\times$MXFP4 & MXFP4$\times$MXFP4 \\
\midrule
$512\times4096\times4096$   & 46.6 & 1.93$\times$ & 0.98$\times$ & 0.78$\times$ & 0.86$\times$ & 0.41$\times$ & 0.30$\times$ \\
$1024\times4096\times4096$  & 46.5 & 1.93$\times$ & 0.97$\times$ & 0.83$\times$ & 0.87$\times$ & 0.45$\times$ & 0.31$\times$ \\
$4096\times4096\times4096$  & 61.9 & 1.97$\times$ & 1.00$\times$ & 0.85$\times$ & 0.92$\times$ & 0.45$\times$ & 0.29$\times$ \\
$2048\times8192\times8192$  & 45.3 & 1.91$\times$ & 0.98$\times$ & 0.85$\times$ & 0.91$\times$ & 0.43$\times$ & 0.29$\times$ \\
$4096\times14336\times4096$ & 54.5 & 1.96$\times$ & 0.97$\times$ & 0.82$\times$ & 0.90$\times$ & 0.46$\times$ & 0.27$\times$ \\
$8192\times8192\times4096$  & 58.2 & 1.96$\times$ & 0.99$\times$ & 0.84$\times$ & 0.92$\times$ & 0.46$\times$ & 0.28$\times$ \\
\bottomrule
\end{tabular}
\end{table}

\section{Kernel details}
\label{app:kernels}

\paragraph{Fused vs.\ two-pass.} A single kernel that unpacks each $64\times128$ weight tile (64 output channels by 128 reduction
elements) into threadgroup memory inside the K-loop and calls \texttt{matmul2d} on it runs at $0.39$--$0.48\times$ FP16 on LLM-sized layers,
although it is still $1.2$--$1.3\times$ faster than the identical kernel with FP16 tiles: every M-tile re-unpacks the
same weights, and the scalar unpack competes with the matrix unit. Unpacking each layer once into an INT8 buffer
(about 1.5 bytes moved per weight; 0.03\,ms for $4096\times4096$, 0.15\,ms for $14336\times4096$) and running a plain
INT8 \texttt{matmul2d} from device memory is $1.36$--$1.51\times$ faster than FP16 with 16-bit weights. In
Table~\ref{tab:layer} this path also includes the activation quantization (0.04\,ms at $2048\times4096$) and the
scaling of the outputs (0.12\,ms at $2048\times4096$ outputs), both linear in the number of tokens; the W8A8 bound
of Table~\ref{tab:layer} includes the activation quantization but neither unpack nor scaling, which is why the gap between them grows with $M$.

\paragraph{Epilogues.} Scales applied through the tensor API's cooperative destination tensor in the GEMM's registers
are about 10\% slower than an INT32 output plus a separate scaling pass on single layers, but in the fused model the
in-register epilogue (scale; scale and residual; SwiGLU on gate/up rows interleaved in pairs) is $1.04$--$1.12\times$ faster at
2048--8192 tokens, where the INT32 round trip of the 12288-wide gate/up output dominates.

\paragraph{MoE.} One counting-sort kernel groups token--expert assignments by expert (replacing two argsorts and
gathers; identical logits) and returns each assignment's position in expert order; the INT8 activation rows are
gathered into that order, one per assignment, and the combine kernel reads each token's top-$k$ output rows back by
position, weights them by the routing scores and adds the residual, fused with the next RMSNorm and activation
quantization. The grouped GEMM is one dispatch over all experts; each threadgroup finds its expert and
row tile from a prefix sum of per-expert tile counts. With in-kernel unpack each threadgroup unpacks its expert's
$64\times64$ weight tile per K-chunk of 64 into threadgroup memory (K-chunk 64 beats 128 at every length: half the
threadgroup memory, twice the occupancy). The two-pass variant unpacks all of a layer's experts into an INT8 temporary of $E\cdot N\cdot K$ bytes
(0.27\,GB for OLMoE's stacked gate/up projections, 0.54\,GB for Qwen3.5-35B-A3B's). Two-pass is $0.82\times$ MLX at
512 tokens on OLMoE and $0.66\times$ on
Qwen3.5-35B-A3B, because unpacking every expert costs more than the few-row GEMMs; from about 256 rows per expert
the two variants are even.

\paragraph{Decode.} The GEMV turns each 4-bit code $c$ (sign and E2M1 index) into the half-precision number
$2v\cdot2^{-15}$ with one bit operation whose 16-bit result is reinterpreted as a half,
$((c\,\&\,7) \ll 9) \mid ((c\,\&\,8) \ll 12)$ (index 1 becomes the subnormal $2^{-15}$), multiplies it by the INT8 activations held as floats and sums each 32-block exactly in FP32 (a
multiple of $2^{-15}$ below 2). It multiplies the block sum by the block multiplier ($2^{e_{n,b}-E_n}$, or $m$ for
the integer-multiplier formats), accumulates a lane's blocks in FP32, which stays exact below $2^{24}$ units of
$2^{-15}$, converts the lane sum exactly to the integer $\sum_k a_{m,k}q_{n,k}$ as $\mathrm{rint}(2^{15}\,\mathrm{acc})$,
reduces across lanes in INT32 and applies the same epilogues. A row's reduction is split over the 32 lanes of a SIMD
group, 32 codes per lane per step, so a lane accumulates $K/32$ products of at most $127\cdot120$ (16-wide format)
or $127\cdot96$ (SC-MXFP4) units of $2^{-15}$, which stays below $2^{24}$ for $K$ up to 35k or 44k (448 products per
lane at $K = 14336$); for MoE models it gathers the selected experts' rows itself. Folding the
activation quantization into the GEMV prologue made decode $0.61$--$0.74\times$ slower with two SIMD groups per threadgroup, because every threadgroup
recomputes the row statistics; for Qwen3.5-35B-A3B's output projections ($2048\times4096$) with 16 SIMD groups per threadgroup (64 threadgroups instead of 256) it saves 0.2\,ms per token against a separate quantization kernel, with identical logits; on Qwen3-1.7B's o and down projections the change is within session noise ($0.98\times$ either way). Table~\ref{tab:system} uses it for both models.

\section{Linear-layer latency}
\label{app:layer}

\begin{table}[h]
  \centering\small
  \caption{Linear-layer latency (ms, median) on M5 Max; all W4 variants except MLX read the same SC-MXFP4 weights.
  Our kernels are timed with GPU timestamps: the elapsed time of one command buffer that runs a pipeline's kernels
  back to back, divided by the number of runs it contains, so gaps between a pipeline's kernels count: \texttt{matmul2d} FP16 is a plain
  FP16 GEMM; W4A16 fused unpacks to FP16 tiles inside the K-loop; W4A16 2-pass unpacks to an FP16 buffer and runs an
  FP16 GEMM with the row scales; W4A8 quantizes the activations, unpacks to an INT8 buffer and runs an INT8 GEMM,
  scaling in registers or in a separate pass, whichever is faster; $^\dagger$W8A8 quantizes the activations and runs
  an INT8 GEMM with INT32 output from 8-bit weights (no unpack, no scaling), a bound on the achievable speed. MLX's
  kernels are timed by wall clock in a separate process (10 calls per evaluation), which includes some host overhead, so the ``mixed'' vs-MLX column mixes two timing methods. ``Alike'': our MLX-integrated kernels (the ones used
  end to end) and MLX's, called from MLX and timed alike by wall clock in one process, interleaved; at $K=14336$ the
  MLX-integrated W4A8 is slower than the standalone one (3.90 vs.\ 2.70\,ms). Like for like are this column and the
  comparisons against \texttt{matmul2d} FP16 and W4A16 2-pass.}
  \label{tab:layer}
  \setlength{\tabcolsep}{3.3pt}\begin{tabular}{lrrrrrrrrr}
\toprule
 & MLX & \texttt{matmul2d} & MLX & W4A16 & W4A16 & \textbf{W4A8} & & \multicolumn{2}{c}{vs MLX MXFP4} \\
\cmidrule(lr){9-10}
$M\times N\times K$ & FP16 & FP16 & MXFP4 & fused & 2-pass & \textbf{(ours)} & W8A8$^\dagger$ & mixed & alike \\
\midrule
512$\times$4096$\times$4096 & 0.321 & 0.300 & 0.346 & 0.934 & 0.361 & \textbf{0.220} & 0.168 & 1.57$\times$ & \textbf{1.48$\times$} \\
1024$\times$4096$\times$4096 & 0.677 & 0.583 & 0.716 & 1.824 & 0.666 & \textbf{0.423} & 0.319 & 1.69$\times$ & \textbf{1.69$\times$} \\
2048$\times$4096$\times$4096 & 1.360 & 1.178 & 1.446 & 3.443 & 1.242 & \textbf{0.838} & 0.628 & 1.72$\times$ & \textbf{1.80$\times$} \\
4096$\times$4096$\times$4096 & 2.957 & 2.273 & 3.212 & 6.330 & 2.420 & \textbf{1.581} & 1.259 & 2.03$\times$ & \textbf{1.86$\times$} \\
2048$\times$14336$\times$4096 & 4.954 & 4.000 & 5.358 & 11.198 & 4.348 & \textbf{2.703} & 1.992 & 1.98$\times$ & \textbf{1.90$\times$} \\
2048$\times$4096$\times$14336 & 6.718 & 4.071 & 5.189 & 13.455 & 4.314 & \textbf{2.696} & 2.374 & 1.92$\times$ & \textbf{1.37$\times$} \\
\bottomrule
\end{tabular}

\end{table}

\section{Measurement protocol}
\label{app:method}

All measurements are on one Apple M5 Max (128\,GB, macOS~26.6.2 unless marked 27.2) on AC power in High Power mode. Kernel
microbenchmarks use GPU timestamps; every repetition times all compared kernels back to back after a 1\,s clock
warm-up, and speedups are medians of per-repetition ratios. End-to-end measurements use wall clock in one process,
with 1\,s warm-up, two warm-up passes per configuration and rotated interleaving of methods and lengths; medians of
7--15 repetitions. The GPU runs in two clock regimes: isolated prompts on an idle GPU (burst) are faster than prompts
following long ones (sustained), and the speedup can differ between the two, so we report both
where they differ (e.g.\ OLMoE prefill at 512 tokens: $1.21\times$ burst, $1.25\times$ sustained). Every timing run
waits until GPU utilization is at most 6\% and starts after 5\,min of idle time; it is rejected if utilization stays above 15\% after it or if MLX MXFP4
deviates from its reference time by more than 7\% (prefill, burst), 10\% (prefill, sustained) or 4\% (decode) (MLX's kernels are the same in every run, so a slower MLX
time flags interference; the check can only remove runs whose baseline was slowed, and in the rejected multi-length Qwen3.5-35B sessions (which bundled 512, 2048, and 8192 tokens together) both MLX and our 512-token runs slowed---ours to $0.6$--$0.7\times$ MLX for reasons not diagnosed---whereas in dedicated single-length 512-token sessions ours reaches $1.16\times$ MLX). OLMoE and the 125B model have no reference values, and the 35B burst reference predates macOS~27.2, on which MLX runs about 11\% faster, so that check is loose. This check was automated only for the final measurements; earlier sessions were re-scored offline against the same references, and the cells from the four sessions that failed were re-timed. Across a configuration's repetitions, the range of prefill times is a median
11\% (ours) and 13\% (MLX) of the median. MLX baselines are timed in the same process as ours; the layer-level MLX comparison
(Appendix~\ref{app:layer}) is timed by wall clock in a separate process. Table~\ref{tab:regimes} lists every
prefill measurement behind the ranges in the text. The default MoE configuration unpacks expert weights inside the
grouped GEMM below 256 rows per expert and in a separate pass from there on; where a forced variant runs the same
kernels as the default (OLMoE: in-kernel at 512 tokens, two-pass from 2048; Qwen3.5: in-kernel up to 2048 tokens,
two-pass at 8192), the medians differ by up to 10\%, the run-to-run spread of these measurements.

\begin{table}[h]
\centering\small
\caption{Prefill time to first token (ms, median of 7--10) by clock regime, with the speedup over MLX MXFP4 in the
same run in parentheses (ratio of medians). MoE: the default configuration (W4A8) and the two expert-weight paths
forced at every length (OLMoE: one session of their own with its own MLX baseline). Qwen3-1.7B and OLMoE: macOS~27.2, default rows the mean of two sessions. -- = not measured. $^\dagger$Re-timed on macOS~27.2 after the layout fix, five methods per process. $^\ddagger$Final protocol (Table~\ref{tab:system}): three methods per process, anchor-checked; 2048/8192 mean of two sessions with both formats loaded, 512 a session of its own.}
\label{tab:regimes}
\setlength{\tabcolsep}{4pt}
\begin{tabular}{llrrrr}
\toprule
Regime & System & 512 & 2048 & 4096 & 8192 \\
\midrule
\multicolumn{6}{l}{\emph{Qwen3-1.7B (dense)}} \\
burst & MLX MXFP4 & 31.4 & 121.0 & -- & -- \\
 & W4A8 & 21.9 (1.43) & 70.5 (1.72) & -- & -- \\
sustained & MLX MXFP4 & 34.0 & 127.4 & -- & 682.3 \\
 & W4A8 & 23.8 (1.43) & 74.6 (1.71) & -- & 406.0 (1.68) \\
\midrule
\multicolumn{6}{l}{\emph{OLMoE-1B-7B (MoE, 64 experts)}} \\
burst & MLX MXFP4 & 35.2 & 111.2 & 223.9 & -- \\
 & W4A8 & 29.2 (1.21) & 83.2 (1.34) & 155.6 (1.44) & -- \\
sustained & MLX MXFP4 & 39.6 & 115.9 & 234.9 & 559.5 \\
 & W4A8 & 31.8 (1.25) & 84.3 (1.37) & 162.7 (1.44) & 350.8 (1.60) \\
forced & MLX MXFP4 & 37.0 & 114.2 & 259.2 & 554.7 \\
(sustained) & in-kernel & 31.8 (1.16) & 97.5 (1.17) & 198.6 (1.30) & 436.9 (1.27) \\
 & two-pass & 44.8 (0.82) & 84.8 (1.35) & 169.4 (1.53) & 354.7 (1.56) \\
\midrule
\multicolumn{6}{l}{\emph{Qwen3.5-35B-A3B (MoE, 256 experts)}} \\
burst & MLX MXFP4 & 126.7 & 370.1 & -- & -- \\
 & W4A8 & 109.9 (1.15) & 300.5 (1.23) & -- & -- \\
 & in-kernel & 109.8 (1.15) & 296.1 (1.25) & -- & -- \\
 & two-pass & 198.8 (0.64) & 361.7 (1.02) & -- & -- \\
sustained & MLX MXFP4 & 134.9 & 386.0 & -- & 1630.6 \\
 & W4A8 & 122.7 (1.10) & 317.5 (1.22) & -- & 1140.2 (1.43) \\
 & in-kernel & 111.5 (1.21) & 306.0 (1.26) & -- & 1183.2 (1.38) \\
 & two-pass & 202.9 (0.66) & 367.2 (1.05) & -- & 1158.0 (1.41) \\
re-timed$^\dagger$ & MLX MXFP4 & 136.5 & 370.6 & -- & 1555.2 \\
 & W4A8 & 116.0 (1.18) & 296.9 (1.25) & -- & 1140.6 (1.36) \\
 & in-kernel & 105.7 (1.29) & 297.2 (1.25) & -- & 1142.1 (1.36) \\
 & two-pass & 198.6 (0.69) & 349.5 (1.06) & -- & 1105.2 (1.41) \\
final$^\ddagger$ & MLX MXFP4 & 113.0 & 380.0 & -- & 1530.0 \\
 & W4A8 & 97.7 (1.16) & 332.6 (1.14) & -- & 1127.9 (1.36) \\
 & 8-wide & 101.1 (1.12) & 322.6 (1.18) & -- & 1144.3 (1.34) \\
\bottomrule
\end{tabular}
\end{table}

\paragraph{Qwen3.8-Flash-Next.} The model code is mlx-lm pull request~1788 at a pinned commit. Weights are read
layer by layer from memory-mapped safetensors of the official FP8 checkpoint (E4M3 decoded exactly with a 256-entry
table, times the per-$128\times128$-tile scales). The 51B-parameter n-gram table is never loaded: the rows a batch
addresses are gathered from the mapped FP8 shards on the host, for every system alike. Our system replaces only the
W4A8 projections: the model's own DeltaNet, attention and residual-mixing code runs with stub modules that return
our GEMMs' outputs, and the output projections' inputs are quantized and multiplied by our kernels, so every
operation between the GEMMs is the reference implementation's. The layer-streamed perplexity keeps, for each method,
the hidden states of all 145 windows (four residual streams, BF16, about 6\,GB per method); a method's weights are
fake-quantized once per layer, experts 64 at a time. The perplexity tables use the model's op-by-op reference
recurrence for Gated DeltaNet; its Metal kernel, which both timed systems use, changes the
perplexity of the first 8 windows by $-0.3$ to $+0.6\%$ and not the order of the formats (FP8: 2.326 vs.\ 2.325;
fake-quantized SC-MXFP4 2.640 vs.\ 2.638, 16-wide 2.514 vs.\ 2.521, MLX MXFP4 W4A16 2.756 vs.\ 2.764; on our
kernels 2.647 vs.\ 2.633 and 2.511 vs.\ 2.511); the GSM8K runs use the Metal kernel. Packing the 125B model takes about 30\,min for SC-MXFP4 and 52\,min for the 16-wide format. The timed
systems need 59 (SC-MXFP4), 62 (16-wide) and 65\,GiB (8-wide; MLX MXFP4 65\,GiB) and are measured in
separate processes, alternating, each session under the protocol above.

\section{Formats per class of matrices in MoE models}
\label{app:scope}

\begin{table}[h]
\centering\small
\caption{Qwen3.5-35B-A3B, WikiText-2 perplexity (145 windows, fake quantization, INT8 activations at every W4A8 matrix) when the routed and shared experts and the dense projections (DeltaNet in\_proj\_qkv / in\_proj\_z / out\_proj, attention q / k / v / o) get different weights. bits/w: over all W4A8 weights, row scales included; GB/step: W4A8 weight bytes one decode step reads (every dense projection, 9 of 257 experts per layer). Intervals: 95\% paired bootstrap against the 8-wide format everywhere. \texttt{results/36\_ppl\_qwen35.csv}, \texttt{36\_bootstrap.csv}; BF16 5.723.}
\label{tab:scope}
\begin{tabular}{llrrrr}
\toprule
Experts & Dense projections & bits/w & GB/step & Perplexity & $-$ 8-wide everywhere \\
\midrule
BF16 & BF16 & 16 & & 5.779 & \\
8-wide & BF16 & & & 5.828 & \\
BF16 & 8-wide & & & 5.839 & \\
SC-MXFP4 & BF16 & & & 5.891 & \\
\midrule
8-wide & 8-wide & 4.53 & 1.36 & \textbf{5.887} & \\
NVFP4, search & NVFP4, search & 4.50 & 1.36 & 5.907 & $+0.020$ $[+0.012,+0.029]$ \\
16-wide & 16-wide & 4.28 & 1.29 & 5.909 & \\
SC-MXFP4 & SC-MXFP4 & 4.09 & 1.23 & 6.051 & \\
\midrule
8-wide & INT8 (per row) & 4.66 & 1.92 & \textbf{5.829} & $-0.058$ $[-0.067,-0.050]$ \\
SC-MXFP4 & INT8 (per row) & 4.24 & 1.86 & 5.892 & $+0.005$ $[-0.003,+0.014]$ \\
16-wide & 8-wide & 4.29 & 1.33 & 5.902 & $+0.015$ $[+0.009,+0.022]$ \\
SC-MXFP4 & 8-wide & 4.11 & 1.30 & 5.950 & $+0.063$ $[+0.055,+0.071]$ \\
\bottomrule
\end{tabular}
\end{table}

\begin{table}[h]
\centering\small
\caption{Qwen3.8-Flash-Next (125B), WikiText-2 perplexity (145 windows, layer-streamed fake quantization, INT8 activations at every W4A8 matrix; FP8 checkpoint 4.662). Columns as in Table~\ref{tab:scope}; GB/step counts 11 of 513 experts per layer. Intervals: 95\% paired bootstrap against the 8-wide format everywhere. \texttt{results/30\_ppl\_qwen38.csv}, \texttt{37\_ppl\_qwen38.csv}, \texttt{37\_bootstrap\_qwen38.csv}; the other rows from Table~\ref{tab:q38}. $^\ast$Routed experts only; the shared expert and the dense projections keep the checkpoint's weights and 16-bit inputs (the scope of the public releases), so this row is not comparable in scope with the others. KL: mean per-token KL divergence from the FP8 model over the first 32 windows (\texttt{40\_kl\_qwen38.csv}, \texttt{11\_kl\_qwen38.csv}).}
\label{tab:scope125}
\begin{tabular}{llrrrrr}
\toprule
Experts & Dense projections & bits/w & GB/step & Perplexity & $-$ 8-wide everywhere & KL \\
\midrule
8-wide, routed only$^\ast$ & as checkpoint & & & 4.658 & & \\
\midrule
8-wide & 8-wide & 4.53 & 2.98 & 4.842 & & 0.145 \\
NVFP4, search & NVFP4, search & 4.50 & 2.96 & 4.917 & $+0.075$ $[+0.056,+0.095]$ & \\
16-wide & 16-wide & 4.27 & 2.81 & 4.969 & & 0.177 \\
SC-MXFP4 & SC-MXFP4 & 4.09 & 2.69 & 5.107 & & 0.257 \\
\midrule
16-wide & INT8 (per row) & 4.36 & 4.06 & \textbf{4.711} & $-0.131$ $[-0.147,-0.115]$ & \textbf{0.071} \\
SC-MXFP4 & INT8 (per row) & 4.17 & 4.00 & 4.746 & $-0.097$ $[-0.115,-0.078]$ & \\
16-wide & 8-wide & 4.28 & 2.89 & 4.852 & $+0.010$ $[-0.001,+0.021]$ & 0.151 \\
SC-MXFP4 & 8-wide & 4.10 & 2.83 & 4.890 & $+0.048$ $[+0.034,+0.062]$ & 0.179 \\
\bottomrule
\end{tabular}
\end{table}

On both models the dense projections are a small share of the weights (3.8\% and 2.2\%) and about half of the bytes a batch-1 decode step reads. Giving them 8 bits removes most of the 4-bit loss at a few hundredths of a bit per weight (on the 125B model it also halves the KL divergence from the FP8 model, 0.071 against 0.145), but raises the bytes per decode step by 40--50\%; it suits prefill-heavy use. With 8-wide dense projections, the experts can drop to 16-wide blocks on the 125B model at no measurable cost, and to SC-MXFP4 at $+0.05$ (125B) to $+0.06$ (35B). Our kernels hold one format per model, so none of these mixed configurations has been timed.

\section{Tokens per step}
\label{app:tokens}

\begin{table}[h]
\centering\small
\caption{Qwen3.5-35B-A3B on an M5 Max (macOS~27.2): time of one forward step of $T$ new tokens on a cache of 512 or 2048 tokens, with the logits of all $T$ positions, synchronized after each step; median of 25 ($T\le32$) or 15 steps, methods interleaved, idle-GPU gate (no MLX reference exists for this protocol, so no anchor check). Ours: $T=1$ decode GEMVs, $T\ge2$ the prefill path; MLX's MXFP4 output projection for both systems. ``vs.\ MLX'': MLX MXFP4 time / ours. MLX's own time is not monotone in $T$ (81\,ms at 96 tokens, 59\,ms at 128), which makes the 48--96 column look better for us than the trend. \texttt{results/37\_tokens\_qwen35\_*.csv} ($T\le32$), \texttt{38\_tokens\_qwen35\_*.csv} ($T\ge48$); \texttt{paper/make\_tokens\_table.py}.}
\label{tab:tokens}
\resizebox{\textwidth}{!}{\begin{tabular}{rrrrrrrrr}
\toprule
T & \multicolumn{2}{c}{8-wide, 512} & \multicolumn{2}{c}{8-wide, 2048} & \multicolumn{2}{c}{SC-MXFP4, 512} & \multicolumn{2}{c}{SC-MXFP4, 2048} \\
 & ms & vs.\ MLX & ms & vs.\ MLX & ms & vs.\ MLX & ms & vs.\ MLX \\
\midrule
1 & 9.0 & 0.93 & 9.2 & 0.94 & 8.7 & 0.91 & 8.9 & 0.96 \\
2 & 21.1 & 0.44 & 21.0 & 0.46 & 20.4 & 0.43 & 20.3 & 0.47 \\
3 & 22.3 & 0.46 & 23.1 & 0.48 & 21.5 & 0.48 & 22.3 & 0.49 \\
4 & 23.7 & 0.51 & 25.0 & 0.50 & 23.1 & 0.52 & 24.7 & 0.51 \\
6 & 26.0 & 0.57 & 27.7 & 0.56 & 25.7 & 0.58 & 27.3 & 0.57 \\
8 & 28.8 & 0.60 & 30.8 & 0.59 & 28.1 & 0.61 & 30.2 & 0.60 \\
12 & 33.3 & 0.67 & 35.0 & 0.67 & 32.3 & 0.69 & 33.6 & 0.69 \\
16 & 34.5 & 0.74 & 36.0 & 0.73 & 32.8 & 0.78 & 34.3 & 0.76 \\
24 & 37.0 & 0.88 & 40.0 & 0.86 & 35.8 & 0.92 & 37.2 & 0.90 \\
32 & 39.2 & 0.93 & 41.5 & 0.91 & 37.9 & 0.96 & 38.7 & 0.95 \\
48 & 44.2 & 1.13 & 47.2 & 1.08 & 42.6 & 1.17 & 45.3 & 1.14 \\
64 & 46.3 & 1.25 & 50.9 & 1.18 & 44.6 & 1.30 & 49.0 & 1.25 \\
96 & 53.0 & 1.53 & 56.5 & 1.49 & 50.5 & 1.60 & 54.4 & 1.56 \\
128 & 57.3 & 1.03 & 63.8 & 0.98 & 55.0 & 1.07 & 61.8 & 1.03 \\
192 & 69.9 & 0.98 & 76.9 & 0.95 & 67.4 & 1.02 & 74.1 & 0.99 \\
256 & 80.9 & 1.02 & 86.4 & 0.99 & 78.9 & 1.04 & 84.8 & 1.02 \\
384 & 97.8 & 1.12 & 106.3 & 1.08 & 95.6 & 1.15 & 104.0 & 1.12 \\
512 & 115.9 & 1.15 & 125.5 & 1.13 & 111.9 & 1.19 & 122.7 & 1.17 \\
\bottomrule
\end{tabular}
}
\end{table}

\section{Accuracy ablations}
\label{app:ablation}

\begin{table}[h]
  \centering\small
  \caption{WikiText-2 perplexity ablations (same protocol as Table~\ref{tab:ppl}). The two W4A16 rows with exponent
  search show three decimals for the comparison in \S\ref{sec:acc}. -- = not run.}
  \label{tab:ablation}
  \setlength{\tabcolsep}{4pt}\begin{tabular}{lrrrrrr}
\toprule
 & \multicolumn{3}{c}{Qwen3} & SmolLM3 & OLMoE & Qwen3.5 \\
\cmidrule(lr){2-4}
Method & 0.6B & 1.7B & 4B & 3B & 1B-7B$^\ddagger$ & 35B-A3B$^\ddagger$ \\
\midrule
BF16 weights, INT8 activations & 12.98 & 9.64 & 8.13 & 8.00 & 7.59 & 5.78 \\
MXFP4, OCP exponent rule, W4A16 & 15.19 & 10.81 & 8.54 & 8.71 & -- & -- \\
MXFP4, exponent search, W4A16 & 15.198 & 10.697 & 8.506 & 8.550 & -- & -- \\
SC-MXFP4 W4A16 & 15.195 & 10.704 & 8.506 & 8.548 & 8.036 & 6.001 \\
SC-MXFP4 + GPTQ W4A16 & -- & 10.22 & -- & -- & -- & -- \\
SC-MXFP4 W4A8 & 15.55 & 10.99 & 8.75 & 8.62 & 8.11 & 6.05 \\
SC-MXFP4 W4A8 + Hadamard (128-blocks) & 17.45 & 13.31 & 8.96 & -- & -- & -- \\
SC-MXFP4 W4A8 + Hadamard (32-blocks) & -- & 13.18 & 8.80 & -- & -- & -- \\
E2M1, 16-blocks, $m\le10$ W4A16 & -- & 10.13 & -- & -- & -- & -- \\
\bottomrule
\end{tabular}

\end{table}

GPTQ (calibrated on WikiText-2 train, non-sequential, exponents fixed) lowers Qwen3-1.7B W4A8 perplexity on the
deployed kernels from 10.99 to 10.53 but lowers the zero-shot mean from 61.9 to 60.6 (ARC-e $-4.1$, ARC-c $-3.3$,
HellaSwag and PIQA $-1.1$, LAMBADA $+1.8$), consistent with overfitting or scale-coarseness sensitivity. That does not hold across formats: in the simulated runs below, the zero-shot mean of the 8-wide format rises from 63.3 to 64.0 and that of the 16-wide format is 62.8 (62.7 on the kernels without GPTQ), indicating that SC-MXFP4's coarser power-of-two block scales make it more sensitive to fixed-scale GPTQ error compensation. Testing out-of-domain calibration on $128\times 2048$ C4 train tokens (\texttt{sim/colab\_j5\_c4\_gptq.py}, simulated) confirms that most of the WikiText-2 perplexity gain was domain-specific---C4 GPTQ improves WikiText-2 perplexity only modestly over RTN ($10.98\to 10.89$, $-0.092\,[-0.131,-0.054]$, vs.\ $10.55$ on WikiText-2 train for SC-MXFP4; $10.27\to 10.24$, $-0.033\,[-0.059,-0.007]$, vs.\ $10.09$ for 8-wide)---while the SC-MXFP4 zero-shot drop persists (61.95\% RTN $\to$ 61.02\% WT2-GPTQ $\to$ 60.95\% C4-GPTQ; 8-wide 63.71\% $\to$ 63.61\% / 63.04\%), supporting calibration-free RTN as our primary configuration. The two paths also differ slightly on the same weights: the simulated 8-wide format scores 63.3 against 63.5 on the kernels, with perplexity within 0.003, so simulated and kernel zero-shot means are not comparable to better than about 0.2.

\begin{table*}[h]
  \centering\small
  \caption{Perplexity with GPTQ and with GPTQ plus rotations (fake quantization, WikiText-2 test; GPTQ calibrated on WikiText-2 train, exponents fixed; Hadamard rotation of each 32-wide input block, before the INT8 step). The ``round'' column is round-to-nearest, from Tables~\ref{tab:ppl} and~\ref{tab:ablation}; the BF16-weights ``+rot.'' value has rotation only. BF16 without INT8 activations: 9.40, 7.90, 7.91.}
  \label{tab:gptq}
  \setlength{\tabcolsep}{4pt}\begin{tabular}{lrrrrrrrrr}
\toprule
 & \multicolumn{3}{c}{Qwen3-1.7B} & \multicolumn{3}{c}{Qwen3-4B} & \multicolumn{3}{c}{SmolLM3-3B} \\
\cmidrule(lr){2-4}\cmidrule(lr){5-7}\cmidrule(lr){8-10}
Weights, INT8 activations & round & GPTQ & +rot. & round & GPTQ & +rot. & round & GPTQ & +rot. \\
\midrule
BF16 weights & 9.64 & -- & 9.44 & 8.13 & -- & 7.95 & 8.00 & -- & 7.92 \\
NVFP4, scale search (4.50 bits) & 10.34 & 10.13 & 9.68 & 8.44 & 8.26 & 8.08 & 8.30 & 8.17 & 8.07 \\
SC-MXFP4 (4.07) & 10.99 & 10.52 & 9.75 & 8.75 & 8.35 & 8.17 & 8.62 & 8.30 & 8.19 \\
E2M1, 16-blocks, $m\le10$ (4.26) & 10.39 & 10.18 & 9.67 & 8.42 & 8.27 & 8.08 & 8.35 & 8.17 & 8.08 \\
E2M1, 8-blocks, $m\le10$ (4.51) & \textbf{10.27} & \textbf{10.08} & \textbf{9.61} & \textbf{8.37} & \textbf{8.23} & \textbf{8.04} & \textbf{8.25} & \textbf{8.14} & \textbf{8.04} \\
\bottomrule
\end{tabular}

\end{table*}

\paragraph{Rotations.} Block-wise Hadamard rotations, which MR-GPTQ~\citep{mrgptq} combines with GPTQ to improve MXFP4,
increase the error of every E2M1-based format and of NVFP4 under round-to-nearest in our measurements
(Table~\ref{tab:formats}), and raise SC-MXFP4 W4A8 perplexity from 10.99 to 13.18 on Qwen3-1.7B and from 8.75 to
8.80 on Qwen3-4B; they help only the uniform INT4 grid. Rotated weights are a poorer match for the E2M1 grid under direct rounding.

With GPTQ, MR-GPTQ's setting, the result reverses (Table~\ref{tab:gptq}): rotations lower perplexity on all three models and all formats, and the spread constraint does not prevent it. Compared with GPTQ alone, the 95\% paired bootstrap interval of the change is $[-0.50,-0.43]$, $[-0.20,-0.17]$ and $[-0.11,-0.09]$ for the 8-wide format on Qwen3-1.7B, 4B and SmolLM3. Part of the gain is not about weights: with BF16 weights, rotations lower the perplexity of INT8 activations from 9.64 to 9.44, 8.13 to 7.95 and 8.00 to 7.92, nearly to BF16. That change is about as large as the 8-wide format's gain from rotations on Qwen3-4B and SmolLM3, and under half of it on Qwen3-1.7B. So INT8 activations do benefit from having outliers spread across a block.

Under the same treatment the 8-wide format stays ahead of NVFP4 with scale search on all three models: by 0.05, 0.04 and 0.03 with GPTQ, and 0.06, 0.04 and 0.03 with GPTQ and rotations (all intervals exclude zero). The 16-wide format is within 0.06 of it. SC-MXFP4 trails in both settings, by 0.08--0.13 with GPTQ and rotations. These are intervals over one test set's windows. With GPTQ and rotations on Qwen3-1.7B, the zero-shot mean is 63.6 for SC-MXFP4 and 64.5 for the 8-wide format (simulated; BF16 64.6). None of these GPTQ-plus-rotation dense models has run on our kernels (for round-to-nearest rotation on the Qwen3.5-35B-A3B kernels, see \S\ref{sec:acc}).

\paragraph{Activation-aware NVFP4 scale search.} Table~\ref{tab:nvfp4_search} isolates stronger E4M3 scale searches for NVFP4 on Qwen3-1.7B and Qwen3-4B (both W4A8, WikiText-2 test, 146 windows; \texttt{results/55\_nvfp4\_actaware.csv}). Among calibration-free searches, neither Four-Over-Six~\citep{fourover6} nor a wider E4M3 search ($\pm 6$ steps around \texttt{amax/6} plus \texttt{amax/4}) matches calibration-free 8-wide RTN: 8-wide stays ahead by $-0.185\,[-0.210,-0.160]$ and $-0.076\,[-0.109,-0.047]$ on Qwen3-1.7B, and by $-0.115\,[-0.133,-0.097]$ and $-0.034\,[-0.049,-0.019]$ on Qwen3-4B (95\% paired bootstrap intervals).

\begin{table}[h]
  \centering\small
  \caption{Stronger NVFP4 E4M3 scale searches against integer-exact 8-wide on Qwen3-1.7B and Qwen3-4B (W4A8, WikiText-2 test perplexity over 146 windows of 2048 tokens; \texttt{results/55\_nvfp4\_actaware.csv}). Activation-aware variants minimize block-Hessian output error $\mathrm{Tr}(\Delta W H \Delta W^\top)$ using $128\times 2048$ WikiText-2 train tokens.}
  \label{tab:nvfp4_search}
  \setlength{\tabcolsep}{5pt}\begin{tabular}{llrr}
\toprule
Method (W4A8 except BF16) & Scale objective / calibration & Qwen3-1.7B & Qwen3-4B \\
\midrule
BF16 & Unquantized reference & 9.396 & 7.899 \\
\midrule
\multicolumn{4}{l}{\emph{Calibration-free (weight MSE, no training data)}} \\
NVFP4 (\texttt{amax/6}) & Analytic E4M3 scale & 10.510 & 8.478 \\
NVFP4 Four-Over-Six & Best of $\{\texttt{amax/6},\texttt{amax/4}\}$ (W-MSE) & 10.450 & 8.480 \\
NVFP4 search ($\pm 2$) & $\pm 2$ E4M3 steps (W-MSE) & 10.334 & 8.429 \\
NVFP4 wide search ($\pm 6 + \texttt{amax/4}$) & $\pm 6$ E4M3 steps $\cup \{\texttt{amax/4}\}$ (W-MSE) & 10.341 & 8.399 \\
\textbf{8-wide RTN (ours)} & Row base + $m\in\{1..10\}$ (W-MSE) & \textbf{10.265} & \textbf{8.365} \\
\midrule
\multicolumn{4}{l}{\emph{Activation-aware ($128\times 2048$ WikiText-2 train block Gram $H=\mathbb{E}[x_b^\top x_b]$)}} \\
NVFP4 \texttt{act-diag} & Wide E4M3 candidates, diagonal $H_{ii}$ & 10.013 & 8.341 \\
NVFP4 \texttt{act-gram} & Wide E4M3 candidates, $16\times 16$ $H$ & 10.025 & \textbf{8.325} \\
\textbf{8-wide \texttt{act-gram} (ours)} & Row base + $m\in\{1..10\}$, $8\times 8$ $H$ & \textbf{9.932} & 8.351 \\
\bottomrule
\end{tabular}

\end{table}

Adding activation-aware Gram calibration ($H = \mathbb{E}[x_b^\top x_b]$ on $128\times 2048$ WikiText-2 train tokens), following output-aware scale searches~\citep{soscale,osfp4}, cuts NVFP4 perplexity by $-0.310\,[-0.341,-0.280]$ on Qwen3-1.7B (to 10.025; diagonal 10.013) and by $-0.104\,[-0.121,-0.088]$ on Qwen3-4B (to 8.325; diagonal 8.341), overtaking uncalibrated 8-wide RTN. When the same $8\times 8$ block-Gram objective chooses the 8-wide row base and integer multipliers, however, the comparison splits 1--1: on Qwen3-1.7B, 8-wide \texttt{act-gram} drops by $-0.333\,[-0.368,-0.301]$ to 9.932 and beats NVFP4 \texttt{act-gram} by $-0.093\,[-0.112,-0.075]$ (and NVFP4 \texttt{act-diag} by $-0.081\,[-0.099,-0.064]$); on Qwen3-4B, 8-wide improves modestly to 8.351 ($-0.014\,[-0.028,-0.000]$) and trails NVFP4 \texttt{act-gram} by $+0.027\,[+0.011,+0.043]$, reflecting shared-row-scale coupling across channels when rounding without GPTQ column compensation.

\paragraph{Diagnosing the $1.36\times$ layer-error gap, multi-octave bases, and folded SmoothQuant.} Evaluating every third layer of Qwen3-1.7B against oracle block scales (\texttt{results/56\_k1\_diag\_136x.csv}) confirms that 16-wide RTN matches oracle \emph{weight} MSE ($0.996\times$) yet has $1.354\times$ higher weighted output MSE ($0.00736$ vs.\ $0.00543$). By projection, \texttt{mlp.down\_proj} accounts for $94.3\%$ of the total excess $\mathrm{tr}(\Delta W H \Delta W^\top)$---dominated by final layer~27 \texttt{mlp.down\_proj} ($1.498\times$)---where unmodeled activation spikes amplify tiny weight-scale differences; minimizing block-Gram error reduces \texttt{down\_proj} excess from $1.477\times$ to $0.107\times$ ($0.278\times$ overall). Next, widening the 8-wide \texttt{act-gram} row-base search from 1 to 2--3 downward octaves saturates immediately ($9.932\to 9.934\to 9.932$ on 1.7B; $8.348\to 8.348\to 8.345$ on 4B, intervals including zero; Table~\ref{tab:k1_sq}), confirming that single-octave clipping is not the bottleneck. Instead, folding diagonal per-channel SmoothQuant~\citep{smoothquant} ($\alpha=0.5$) offline into preceding RMSNorm weights and up/gate projections leaves BF16 unchanged within $0.002$ perplexity at zero inference FLOPs/latency, cuts pure W16A8 activation loss ($-0.145$ on 1.7B, $-0.117$ on 4B, $-0.015$ on SmolLM3), and preserves block structure. Paired with folded SmoothQuant, calibrated 8-wide \texttt{act-gram} flips Qwen3-4B ($-0.049\,[-0.062,-0.036]$) and beats calibrated NVFP4 on all three dense models ($9.768$ vs.\ $9.802$, $-0.034\,[-0.048,-0.019]$ on 1.7B; $8.167$ vs.\ $8.216$ on 4B; $8.191$ vs.\ $8.202$, $-0.011\,[-0.022,-0.001]$ on SmolLM3) as well as unfolded W4A16 NVFP4 ($-0.470$, $-0.095$, and $-0.097$, all intervals excluding zero). SmoothQuant+RTN alone also beats SmoothQuant+NVFP4 search on all four dense models, including Mistral-7B-v0.3 ($5.430$ vs.\ $5.439$, $-0.009\,[-0.014,-0.004]$, surpassing W4A16 NVFP4 at $5.451$). On Qwen3-1.7B RTN, $\alpha=0.7$ ($9.900$) improves slightly over $\alpha=0.5$ ($9.908$) and $\alpha=0.3$ ($10.048$), while smoothing \texttt{down\_proj} alone recovers $10.051$ ($>59\%$ of the total RTN gain).

\begin{table}[h]
  \centering\small
  \caption{Multi-octave row-base search and zero-overhead folded SmoothQuant ($\alpha=0.5$) across dense models (WikiText-2 test perplexity over 141--163 windows of 2048 tokens; \texttt{results/56\_k1\_hypotheses.csv}, \texttt{56\_bootstrap\_k1.csv}, and \texttt{57\_mistral7b\_ppl.csv}). All SmoothQuant scales fold offline with zero runtime overhead.}
  \label{tab:k1_sq}
  \setlength{\tabcolsep}{4pt}\resizebox{\linewidth}{!}{\begin{tabular}{llrrrr}
\toprule
Method / calibration & Acts & Qwen3-1.7B & Qwen3-4B & SmolLM3-3B & Mistral-7B \\
\midrule
\multicolumn{6}{l}{\emph{Multi-octave row-base search (block-Gram \texttt{act-gram}, $128\times 2048$ WikiText-2 train)}} \\
NVFP4 \texttt{act-gram} & INT8 & 10.019 & \textbf{8.328} & -- & -- \\
8-wide \texttt{act-gram}, 1 octave & INT8 & 9.932 & 8.348 & -- & -- \\
8-wide \texttt{act-gram}, 2 octaves & INT8 & 9.934 & 8.348 & -- & -- \\
8-wide \texttt{act-gram}, 3 octaves & INT8 & \textbf{9.932} & 8.345 & -- & -- \\
\midrule
\multicolumn{6}{l}{\emph{Zero-overhead folded SmoothQuant ($\alpha=0.5$; folded into preceding norm / up projections)}} \\
BF16 (orig / SmoothQuant) & BF16 & 9.396 / 9.396 & 7.898 / 7.900 & 7.906 / 7.907 & 5.318 / 5.319 \\
BF16 weights W16A8 (orig $\to$ SQ) & INT8 & 9.638 $\to$ 9.493 & 8.123 $\to$ 8.006 & 7.994 $\to$ 7.979 & -- \\
NVFP4 W4A16 (unfolded ref.) & BF16 & 10.239 & 8.262 & 8.287 & 5.451 \\
8-wide RTN (unfolded) & INT8 & 10.269 & 8.372 & 8.248 & 5.480 \\
NVFP4 search + SmoothQuant & INT8 & 9.956 & 8.291 & 8.272 & 5.439 \\
\textbf{8-wide RTN + SmoothQuant} & INT8 & \textbf{9.908} & \textbf{8.254} & \textbf{8.209} & \textbf{5.430} \\
NVFP4 \texttt{act-gram} + SmoothQuant & INT8 & 9.802 & 8.216 & 8.202 & -- \\
\textbf{8-wide \texttt{act-gram} + SmoothQuant} & INT8 & \textbf{9.768} & \textbf{8.167} & \textbf{8.191} & -- \\
\bottomrule
\end{tabular}
}
\end{table}

\paragraph{Why rotations help INT8 activations.} Symmetric per-token INT8 sets one step from the token's largest coordinate. Real activations contain single-coordinate spikes: one down-projection input has a coordinate of 14400 in a block whose other coordinates are below 1. The following bound shows how far a rotation can reduce that step.

\begin{proposition}[Rotated INT8 step]
\label{prop:rot}
Let $x\in\mathbb{R}^n$ have blocks $x_1,\ldots,x_B$ of width $b$, a power of two, and let $M=\max_j\|x_j\|_2$. Let $R=(H_b\oplus\cdots\oplus H_b)D$, with $H_b$ the normalized Hadamard matrix and $D$ a diagonal of independent uniform random signs, and let $\Delta(z)=\|z\|_\infty/127$ be the INT8 step of $z$. Then $\|Rx\|_\infty\ge M/\sqrt b$ always, and for $\delta\in(0,1)$, with probability at least $1-\delta$ over $D$,
\[
\|Rx\|_\infty\le M\sqrt{2\ln(2n/\delta)/b}.
\]
Hence a rotation divides the step by at most $\sqrt b$. If the rounding error of $z$ has mean zero and covariance $\Delta(z)^2I/12$, then for any $W$, with $Rx$ multiplied by $WR^\top$ so that the exact product is unchanged, the expected squared output error is $\|W\|_F^2\Delta(Rx)^2/12$, so the rotation divides it by $(\|x\|_\infty/\|Rx\|_\infty)^2\le b$.
\end{proposition}

\begin{proof}
A coordinate of $Rx$ in block $j$ is $b^{-1/2}\sum_k\pm s_kx_{jk}$ with independent signs $s_k$. Hoeffding's inequality gives probability at most $2\exp(-t^2/(2\|x_j\|_2^2))\le\delta/n$ that the sum exceeds $t=M\sqrt{2\ln(2n/\delta)}$; take the union over $n$ coordinates. For the lower bound, $R$ is orthogonal, so $\|x_j\|_2=\|(Rx)_j\|_2\le\sqrt b\,\|Rx\|_\infty$. Since $\|x\|_\infty\le M$, the step falls by at most $\sqrt b$. For the output error, $\mathbb{E}\|WR^\top e\|_2^2=\Delta^2\|WR^\top\|_F^2/12=\Delta^2\|W\|_F^2/12$.
\end{proof}

A block with one spike reaches the lower bound, which is the best case. The high-probability bound guarantees a gain only when $b>2\ln(2n/\delta)$ in that case, about 26 for $n=2048$, $\delta=0.01$, and it is loose in practice: on 4096 sampled tokens per layer of each of the three models with BF16 weights, $\|Rx\|_\infty$ averages 1.25--2.72 times the lower bound, against an upper limit of about 5.1, and no token exceeds that limit. About 0.002\% of tokens at $b=8$ fall slightly below the lower bound in float32; the six we recomputed in float64 satisfy it.

The white-noise model predicts each layer's measured output error to within 0.2\% in the median layer. It fails in a few layers, by more than 10\% in 10--15 layers without rotation and 2--4 at $b=32$, out of 196--252, carrying at most 13\% and 4\% of the summed relative error. The largest misses are down-projections, where the predicted error is far too high, consistent with rare tokens whose step exceeds nearly every coordinate, so that the error is the coordinate itself; there $|e_i|\le\min(|x_i|,\Delta/2)$ is the bound that applies. We did not explain the misses in the other direction (down to a factor of 0.10).

Table~\ref{tab:rotbound} compares the measured output-error gain with the gain in the perplexity gap to BF16, $\ln(\mathrm{ppl}_0/\mathrm{ppl}_{\mathrm{BF16}})/\ln(\mathrm{ppl}_b/\mathrm{ppl}_{\mathrm{BF16}})$, with BF16 weights. Predictions for $b\ne32$ were written before those runs. Rotation without INT8 stays within 0.004 of BF16 at every $b$, so the rotation adds no measurable noise of its own. The gains rise with $b$, apart from one small reversal on SmolLM3, and Qwen3-4B's flatter curve appears in both, but this is an empirical trend and not a consequence of the proposition: across the 15 cells the perplexity gain is 0.46--1.47 times the error gain. On Qwen3-1.7B at $b\le32$, perplexity improves roughly half as much as single-layer output MSE predicts; applying INT8 to one kind at a time sums to 96\% of the full gap with down-projections carrying 62\%, and down-projections alone show the same ratio (gain 5.6 against an error gain of 14.6 at $b=32$), pointing to cross-layer error interactions beyond single-layer MSE.

\begin{table*}[h]
  \centering\small
  \caption{Rotation gains with BF16 weights and INT8 activations (fake quantization; output error on 4096 sampled tokens per layer of a clean forward pass, summed relative error over layers; perplexity on WikiText-2 test). Crest factor$^2$ is $n\|x\|_\infty^2/\|x\|_2^2$, mean over tokens and layers.}
  \label{tab:rotbound}
  \setlength{\tabcolsep}{4pt}\begin{tabular}{lrrrrrrrrrrrr}
\toprule
 & \multicolumn{2}{c}{Crest factor$^2$} & \multicolumn{5}{c}{Output-error gain, block $b$} & \multicolumn{5}{c}{Perplexity-gap gain, block $b$} \\
Model & $b{=}0$ & $32$ & 8 & 16 & 32 & 64 & 128 & 8 & 16 & 32 & 64 & 128 \\
\midrule
Qwen3-1.7B & 404 & 29 & 5.5 & 8.6 & 12.8 & 16.4 & 19.0 & 2.6 & 4.0 & 6.1 & 9.4 & 19.7 \\
Qwen3-4B & 507 & 100 & 3.2 & 4.4 & 5.0 & 5.7 & 7.7 & 2.7 & 4.2 & 4.3 & 5.0 & 8.0 \\
SmolLM3 & 210 & 23 & 4.6 & 6.4 & 8.1 & 9.3 & 10.1 & 6.3 & 9.4 & 8.8 & 12.5 & 14.8 \\
\bottomrule
\end{tabular}

\end{table*}

\paragraph{Where the activation loss enters, and an outlier bypass.} Table~\ref{tab:bypass} quantizes activations at one class of sites at a time (BF16 weights) and then keeps the $k$ largest-magnitude inputs of every token in float at every site, quantizing the rest to INT8 with a scale taken from the rest. Unlike a rotation, this leaves the weight basis alone, so it composes with round-to-nearest weights; in a kernel it is a rank-$k$ float correction per token, which we have not implemented.

\begin{table}[h]
\centering\small
\caption{WikiText-2 perplexity, fake quantization (146 windows). Top: INT8 activations at the input of one class of linear layers only, BF16 weights. Bottom: INT8 activations everywhere with the $k$ largest $|x|$ of each token in float. Rotation and GPTQ rows from Table~\ref{tab:gptq}. \texttt{results/24\_ppl.csv}, \texttt{37\_ppl\_dense.csv}.}
\label{tab:bypass}
\begin{tabular}{lrr}
\toprule
 & Qwen3-1.7B & Qwen3-4B \\
\midrule
BF16 & 9.40 & 7.90 \\
INT8 at q / k / v / o only & 9.40--9.41 & 7.90--7.91 \\
INT8 at gate / up only & 9.42 / 9.43 & 7.94 / 7.93 \\
INT8 at down only & 9.55 & 8.02 \\
INT8 everywhere & 9.64 & 8.13 \\
\quad $k=1$ / $2$ / $4$ in float & 9.54 / 9.47 / 9.44 & 8.00 / 7.95 / 7.93 \\
\quad block rotation ($b=32$) & 9.44 & 7.95 \\
\midrule
8-wide weights, INT8 everywhere & 10.27 & 8.37 \\
\quad $k=2$ / $4$ in float & 10.11 / \textbf{10.07} & 8.22 / \textbf{8.20} \\
\quad GPTQ & 10.08 & 8.23 \\
\quad GPTQ + rotation & 9.61 & 8.04 \\
\bottomrule
\end{tabular}
\end{table}

\section{The integer-exact formats on the kernels}
\label{app:fmtsys}

\begin{table}[h]
\centering
\small
\caption{SC-MXFP4 and the 16-wide and 8-wide integer-multiplier formats on the same INT8 kernels (Qwen3-1.7B, M5 Max,
macOS~27.2, the two sessions of Table~\ref{tab:system}, mean, all four systems in each; prefill at 512 and 2048 tokens in the burst regime, at 8192 in the
sustained regime; decode in one process per format with MLX MXFP4 and SC-MXFP4, whose values are the mean of the two; Table~\ref{tab:system} reports a separate decode session). Perplexity: WikiText-2, W4A8, measured on the kernels; NVFP4 is fake-quantized (it has no exact INT8
execution; $^\ast$with scale search 10.34). Tasks: mean zero-shot accuracy over six lm-evaluation-harness tasks, on the
kernels. bits/w: average over this model's quantized matrices.}
\label{tab:fmtsys}
\begin{tabular}{lrrrrrrr}
\toprule
 & bits/w & PPL & Tasks & \multicolumn{3}{c}{Prefill TTFT (ms)} & Decode \\
 & & & mean & 512 & 2048 & 8192 & tok/s \\
\midrule
MLX MXFP4 (W4A16)        & 4.25 & --    & 61.8  & 31.4 & 121.0 & 682.3 & 364 \\
NVFP4 (W4A8, fake quant)$^\ast$ & 4.50 & 10.51 & --    & --   & --    & --    & --  \\
SC-MXFP4 (W4A8)          & 4.08 & 10.99 & 61.9  & 21.9 & 70.5  & 406.0 & 356 \\
16-wide, $m\le10$ (W4A8) & 4.26 & 10.39 & 62.7 & 22.2 & 70.6 & 397.4 & 342 \\
\textbf{8-wide, $m\le10$ (W4A8)} & 4.51 & \textbf{10.27} & \textbf{63.5} & 22.3 & 70.7 & \textbf{397.2} & 340 \\
\bottomrule
\end{tabular}
\end{table}

The kernels never assume MXFP4: they read a weight as a 4-bit code times an integer block multiplier from a table,
times one FP32 scale per row. Switching to 16-wide blocks with multipliers $1,\dots,10$ (4-bit block codes,
4.26 bits per weight; $|q|\le 120$) therefore changes only the packer. On Qwen3-1.7B the packed weights equal the
fake-quantized ones, all integer accumulations are exact, decode is bit-identical to prefill, and WikiText-2
perplexity on the kernels is 10.387 (reference 10.394) versus 10.985 for SC-MXFP4 and 10.508 for NVFP4 at the same
INT8 activations, which removes 38\% of SC-MXFP4's perplexity loss. Speed is unchanged:
in the two sessions of Table~\ref{tab:system}, prefill takes 22.2, 70.6 and 397\,ms at 512, 2048 and 8192 tokens (SC-MXFP4: 21.9, 70.5, 406;
MLX MXFP4: 31.4, 121.0, 682), i.e.\ $1.42$--$1.72\times$ faster than MLX MXFP4, and decode runs at 342 vs.\
355\,tokens/s (MLX: 361 in the same process; Table~\ref{tab:fmtsys} reports the two-process means). On the 125B MoE of \S\ref{sec:q38} the prefill gain stays ($1.20$--$1.34\times$
vs.\ $1.22$--$1.35\times$ for SC-MXFP4) and decode runs at $1.01\times$ MLX's speed (Table~\ref{tab:q38speed}). Zero-shot accuracy on the kernels rises to a mean of 62.7 over six
tasks (SC-MXFP4: 61.9; MLX MXFP4 W4A16 on the same matrices: 61.8; BF16: 64.5), with significant gains on
LAMBADA (+2.4) and HellaSwag (+1.4) (paired, McNemar $p<10^{-4}$) and no task worse by more than its standard error; the format needs no
calibration data.

The 8-wide format ($B=8$, 4-bit block codes, 4.51 bits per weight) was run the same way on macOS~27.2 (\texttt{e2e/queue\_format.sh}): all 1,409,286,144 weights equal the fake-quantized ones, the integer accumulations checked are exact, and none of 9.7M decode logits differs from one-token-at-a-time prefill. Perplexity on the kernels is 10.271 against 10.268 for fake quantization, and a few activation codes differ between the kernels' and the reference's quantizer (rounding ties). In the same two sessions, prefill takes 22.3, 70.7 and 397\,ms at 512, 2048 and 8192 tokens, and decode runs at 340 vs.\ 357 (SC-MXFP4) and 367\,tokens/s (MLX MXFP4) in one process. Zero-shot accuracy reaches 73.0, 44.8, 65.2, 74.6, 63.4 and 60.2 on the six tasks of Table~\ref{tab:tasks} (mean 63.5).

\section{Zero-shot paired tests and KL divergence from BF16}
\label{app:kl}

Zero-shot (Table~\ref{tab:tasks}), paired over the same examples (mean over the six tasks, bootstrap 95\% intervals). On Qwen3-1.7B the 8-wide format is ahead of MXFP4 W4A16 by 1.8 $[1.1, 2.4]$, of the 16-wide format by 0.9 $[0.2, 1.5]$ and of MLX's default conversion by 2.9 $[2.2, 3.6]$, and 1.0 $[0.4, 1.6]$ below BF16; without LAMBADA its lead over MXFP4 W4A16 is $+0.5$ $[-0.2, 1.3]$. On Qwen3.5-35B-A3B it is 0.9 $[0.3, 1.6]$ below MXFP4 W4A16, tied with the 16-wide format ($-0.4$ $[-0.9, 0.2]$) and MLX's default conversion ($-0.6$ $[-1.3, 0.1]$) and 1.7 $[1.2, 2.3]$ below BF16. Its gap to SC-MXFP4 and MXFP4 W4A16 there is concentrated in the two ARC tasks (ARC-Challenge $-2.8$ and $-2.2$, ARC-Easy $-0.9$ and $-1.2$); on the other four tasks it is $-0.15$ and $-0.55$, with intervals that include zero. On both ARC tasks the other 4-bit systems shift the margin between the gold answer and the distractors above BF16's and the 8-wide format slightly below it, on both models.

\begin{table}[h]
\centering
\small
\caption{Per-token KL divergence of each deployed system's next-token distribution from BF16's, and the share of tokens whose
top-1 prediction differs from BF16's (\%), on the LAMBADA test passages (outside WikiText), first 32 windows of 2048
tokens; the systems of Table~\ref{tab:tasks}, every W4A8 pass on our kernels. BF16 against itself gives 0.}
\label{tab:kl}
\begin{tabular}{lrrrr}
\toprule
 & \multicolumn{2}{c}{Qwen3-1.7B} & \multicolumn{2}{c}{Qwen3.5-35B-A3B} \\
 & KL & top-1 & KL & top-1 \\
\midrule
MXFP4 W4A16 (default) & 0.399 & 42.0 & 0.079 & 18.3 \\
MXFP4 W4A16           & 0.199 & 28.2 & 0.046 & 12.6 \\
W4A8, SC-MXFP4        & 0.188 & 28.2 & 0.049 & 13.2 \\
W4A8, 16-wide         & 0.123 & 22.2 & 0.032 & 10.5 \\
\textbf{W4A8, 8-wide} & \textbf{0.110} & \textbf{22.1} & \textbf{0.029} & \textbf{10.0} \\
\bottomrule
\end{tabular}
\end{table}

The 8-wide format's KL is 41--45\% below SC-MXFP4's and MXFP4 W4A16's on Qwen3-1.7B and 36--40\% on Qwen3.5-35B-A3B. Paired over the 32 windows (bootstrap 95\% intervals), its KL is below the 16-wide format's by 0.0133
$[0.0123, 0.0143]$ on Qwen3-1.7B and 0.0026 $[0.0020, 0.0032]$ on Qwen3.5-35B-A3B, and below SC-MXFP4's by 0.079 and
0.020. On Qwen3.5-35B-A3B SC-MXFP4 is slightly further from BF16 than MLX MXFP4 on the same matrices (0.049 vs.\ 0.046),
the reverse of Qwen3-1.7B.

\section{Spread statistics}
\label{app:spread}

On Qwen3-4B (2.1\% of rows violating), violations concentrate in \texttt{down\_proj} (9.0\% of its rows; 82\% of those caused by a near-zero block,
i.e.\ dead intermediate channels) and \texttt{gate\_proj} (3.6\%; all but one caused by an outlier block, i.e.\
outlier input channels), and stay below 1.3\% elsewhere.

\begin{figure}[h]
  \centering
  \includegraphics[width=0.75\linewidth]{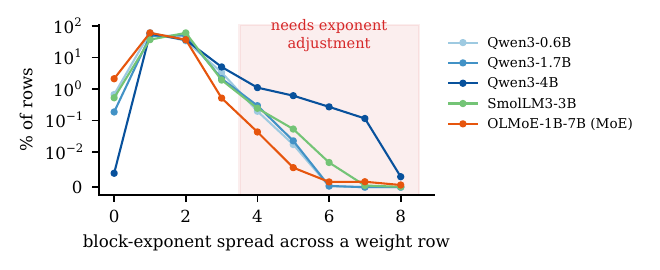}
  \caption{Block-exponent spread per weight row. Spread above 3 (shaded) needs an exponent adjustment. Logarithmic axis; its bottom tick marks 0 (no rows).}
  \label{fig:spread}
\end{figure}

\begin{table}[h]
\centering\small
\caption{Rows whose block exponents (MXFP4 with exponent search) span at most 3 over the whole row, and relative
weight MSE ($\times10^{3}$) before and after enforcing the constraint, over all transformer-block linear layers.
$^\dagger$Every W4A8 matrix of the FP8 checkpoint (95.4M rows; 73,728 of the 74,028 matrices are routed experts),
with the FP8 experts decoded exactly.}
\label{tab:spread}
\begin{tabular}{lrrrr}
\toprule
Model & Matrices & $s_n\le3$ & Max $s_n$ & MSE, MXFP4 $\to$ SC \\
\midrule
Qwen3-0.6B & 196 & 99.8\% & 6 & $12.70 \to 12.70$ \\
Qwen3-1.7B & 196 & 99.7\% & 6 & $12.72 \to 12.72$ \\
Qwen3-4B & 252 & 97.9\% & 8 & $12.74 \to 12.74$ \\
SmolLM3-3B & 252 & 99.7\% & 8 & $12.65 \to 12.66$ \\
Mistral-7B-v0.3 & 224 & 99.5\% & 7 & $12.77 \to 12.77$ \\
OLMoE-1B-7B (MoE) & 3136 & 99.9\% & 8 & $12.53 \to 12.53$ \\
Qwen3.5-35B-A3B (MoE) & 30970 & 99.97\% & 7 & $12.71 \to 12.71$ \\
Qwen3.8-Flash-Next (MoE)$^\dagger$ & 74028 & 99.93\% & 9 & $12.83 \to 12.84$ \\
\bottomrule
\end{tabular}
\end{table}

For reference, on the seven smaller models the relative weight MSE of NVFP4 is $8.99$--$9.02\times10^{-3}$ and of MLX's
affine INT4 (group 64) $8.19$--$8.43\times10^{-3}$, against $12.5$--$12.8\times10^{-3}$ for MXFP4 and SC-MXFP4
(Table~\ref{tab:spread}). On Qwen3.5-35B-A3B, 1.1\% of routed-expert gate/up rows are numerically dead
($|w|\approx10^{-38}$); their row scale is a float32 subnormal that the GPU flushes to zero, so they contribute exactly
0 instead of $\sim10^{-37}$, far below BF16 resolution.

\section{Comparison with llama.cpp's 4-bit types}
\label{app:llamacpp}

Table~\ref{tab:llamacpp} compares llama.cpp's 4-bit types with ours under one perplexity protocol. llama.cpp's
Q4\_K\_M (a k-quant that keeps some tensors at 6 bits) and IQ4\_XS (non-uniform 4-bit levels with a 6-bit scale per
32 weights; IQ4\_KS, the type closest to our format family, is only in the ik\_llama.cpp fork) are quantized from the
BF16 GGUFs without an importance matrix, like every format in this paper, and scored with \texttt{llama-perplexity},
which splits the text into disjoint 2048-token windows and scores only the second half of each. We score our
fake-quantized formats the same way (same text, windows and scored tokens); our BF16 is within 0.11\% of llama.cpp's
on all three models, so each format's change against its own BF16 compares across the two. The coverage still
differs: llama.cpp's types also quantize the token embedding (4--6 bits; MXFP4\_MOE: 8) and keep the output layer at
6--8 bits, Q4\_K\_M and IQ4\_XS keep some attention and FFN matrices at 5--6 bits, and MXFP4\_MOE keeps every matrix
outside the routed experts at 8 bits, whereas ours quantize every attention and MLP matrix to 4 bits and keep the
embedding and output layer in BF16.

With 16-bit activations, as llama.cpp's Metal kernels compute, SC-MXFP4 loses slightly less than Q4\_0 on the dense
Qwen3-1.7B (13.1 vs.\ 14.9\%) but more than every llama.cpp type on the two MoEs (6.9 and 4.9\% against
1.7--4.3\%), consistent with standard MXFP4's power-of-two block scales being coarser than Q4\_0's 16-bit ones and
with MXFP4\_MOE keeping attention at 8 bits. The integer-exact 16-wide format loses less than Q4\_0 on both models
where we ran it, less than IQ4\_XS on Qwen3-1.7B and less than MXFP4\_MOE on OLMoE, with every attention matrix at
4 bits; Q4\_K\_M, at about 5 bits per weight, is more accurate than our formats throughout. INT8 activations, which
our kernels use, add 0.7--2.8 points. Q4\_K\_M and IQ4\_XS prefill more slowly than Q4\_0, so the prefill speedups over
llama.cpp in Table~\ref{tab:system}, which use Q4\_0, are the smallest over any of its 4-bit types we measured: its MXFP4\_MOE type is 11--29\%
slower than Q4\_0 on the two MoEs, its Q4\_K\_M and IQ4\_XS prefill at $0.82$--$0.96\times$ Q4\_0's speed, and every prefill
ratio against Q4\_0 is at least $1.14\times$.

\begin{table}[h]
\centering\small
\caption{llama.cpp's 4-bit types and ours, scored the same way: \texttt{llama-perplexity}'s protocol on the
WikiText-2 test set (disjoint 2048-token windows, second half of each scored; not comparable with
Table~\ref{tab:ppl}), each against its own BF16. llama.cpp: commit f74e2fa, Metal. Ours: fake-quantized as in
Table~\ref{tab:ppl} (W4A8 is what our kernels compute), about 4.1 (SC-MXFP4) and 4.3 (16-wide) bits per weight in the
quantized matrices, embedding and output layer in BF16. Speed: relative to Q4\_0 in the same session, prefill at 512,
2048 and 8192 tokens (OLMoE: 4096; the faster of two batch sizes) and decode at a 512-token context; MXFP4\_MOE's speed
is in Table~\ref{tab:system}. -- = not run or not applicable.}
\label{tab:llamacpp}
\setlength{\tabcolsep}{4pt}\begin{tabular}{lrrr}
\toprule
 & Qwen3-1.7B & OLMoE-1B-7B & Qwen3.5-35B-A3B \\
\midrule
\multicolumn{4}{l}{\emph{BF16 perplexity, llama.cpp / ours}} \\
BF16 & 8.436 / 8.445 & 6.724 / 6.727 & 5.131 / 5.133 \\
\midrule
\multicolumn{4}{l}{\emph{llama.cpp: change against its BF16 (bits per weight of the file)}} \\
Q4\_0 & $+14.9\%$ (4.90) & $+4.3\%$ (4.54) & $+3.2\%$ (4.55) \\
Q4\_K\_M & $+4.8\%$ (5.15) & $+2.2\%$ (4.87) & $+1.9\%$ (4.89) \\
IQ4\_XS & $+8.9\%$ (4.73) & $+2.7\%$ (4.34) & $+1.7\%$ (4.37) \\
MXFP4\_MOE & -- & $+4.2\%$ (4.55) & $+1.8\%$ (4.57) \\
\midrule
\multicolumn{4}{l}{\emph{Ours: change against our BF16}} \\
SC-MXFP4 W4A16 & $+13.1\%$ & $+6.9\%$ & $+4.9\%$ \\
SC-MXFP4 W4A8 & $+15.9\%$ & $+8.3\%$ & $+5.6\%$ \\
16-block, $m\le10$ W4A16 & $+7.7\%$ & $+3.8\%$ & -- \\
16-block, $m\le10$ W4A8 & $+10.4\%$ & $+5.1\%$ & -- \\
\midrule
\multicolumn{4}{l}{\emph{llama.cpp speed relative to Q4\_0: prefill (range over lengths) / decode}} \\
Q4\_K\_M & 0.90--0.95 / 0.95 & 0.87--0.89 / 0.96 & 0.91--0.92 / 0.89 \\
IQ4\_XS & 0.82--0.96 / 0.98 & 0.84--0.90 / 1.00 & 0.92 / 0.92 \\
\bottomrule
\end{tabular}

\end{table}

\section{Qwen3.8-Flash-Next: speed details and concurrent kernels}
\label{app:q38speed}

\paragraph{Accuracy details.} KL divergence from the FP8 model with only the routed experts quantized (32 windows; 16-bit / INT8 activations): $0.0534$ / $0.0557$ for the 16-wide format, $0.0539$ / $0.0553$ for NVFP4 with scale search, $0.0610$ / $0.0628$ for affine INT4, $0.0796$ / $0.0802$ for SC-MXFP4 and $0.0931$ / $0.0940$ for MXFP4 with the OCP rule, the format with the lowest perplexity. The rate of tokens whose top-1 prediction changes orders them the same way, with the 16-wide format and NVFP4 within 0.2 points (7.7--7.9\%). Quantizing the dense projections as well triples the divergence ($0.177$ for the 16-wide format); the public NVFP4 and FP8 releases leave them at higher precision. SC-MXFP4 is more accurate than MXFP4 as MLX quantizes it, because exponent search matters more than the spread constraint. Of the expert rows, 0.004--0.087\% violate the spread condition (the expert down projections have only 20 blocks per row). With SC-MXFP4 weights, INT8 activations change nothing measurable on GSM8K (McNemar $p=0.87$, against MLX's kernels on the same weights) whereas the weight quantizer does ($p=0.010$); the 16-wide format solves more problems than SC-MXFP4 ($p=0.001$). With the 8-wide format (whose pack is within $0.3\%$ of the reference on 8 windows), our system solves $95.30\pm0.58\%$ of GSM8K, significantly beating MLX's MXFP4 ($p<10^{-9}$) and SC-MXFP4 ($p=1.0\times10^{-4}$), and numerically edging 16-wide ($+0.46$ points, $p=0.43$).

\begin{table}[h]
\centering
\small
\caption{Qwen3.8-Flash-Next on an M5 Max, macOS~27.2: time to first token (ms) and batch-1 decode (tokens/s) on exactly the same
matrices. Two sessions per system (one process per system and session), each after a 5\,min pause, mean. Across the two sessions our prefill times agree within 2.4\% and MLX's within 8\%; decode was faster in the second session for every system (MLX 39.7 and 43.0\,tokens/s, ours by 3--6\%). $^\ast$Dense projections decode with MLX's W4A16 GEMV on the same bytes, experts with ours; the 16-wide and 8-wide rows are all ours.}
\label{tab:q38speed}
\setlength{\tabcolsep}{5pt}
\begin{tabular}{lrrrr}
\toprule
 & 512 & 2048 & 8192 & decode \\
\midrule
MLX MXFP4 (W4A16)  & 335 & 1077 & 4919 & 41.4 \\
\textbf{W4A8, SC-MXFP4 (ours)} & \textbf{248} & \textbf{825} & \textbf{4049} & 42.6$^\ast$ \\
W4A8, 16-wide (ours) & 250 & 847 & 4104 & 41.8 \\
W4A8, 8-wide (ours) & 253 & 840 & 4183 & 40.4 \\
\midrule
SC-MXFP4 vs.\ MLX  & $1.35\times$ & $1.30\times$ & $1.22\times$ & $1.03\times^\ast$ \\
16-wide vs.\ MLX  & $1.34\times$ & $1.27\times$ & $1.20\times$ & $1.01\times$ \\
8-wide vs.\ MLX  & $1.32\times$ & $1.28\times$ & $1.18\times$ & $0.98\times$ \\
\bottomrule
\end{tabular}
\end{table}

\paragraph{Speed.} Table~\ref{tab:q38speed} details the 125B timings. In isolation our MoE blocks (router, routed and shared experts) at this model's
shapes, where an expert sees only 10, 40 or 160 rows of a 512-, 2048- or 8192-token prompt, are $1.33$--$1.65\times$
faster than MLX's (random weights, chained calls). End to end the gain is $1.22$--$1.35\times$ (SC-MXFP4): at 2048 tokens the difference
(251\,ms) is mostly the experts' difference ($48\times4.6$\,ms), and about 0.5\,s of each prefill is work outside the W4A8
matrices that both systems share (the DeltaNet recurrence, attention, 96 residual-stream mixers, the n-gram lookup on
the host), a larger share at 8192 tokens. MLX's affine INT4 (68.6\,GiB) is no faster than its MXFP4: in earlier
sessions its prefill is within about 2\% and its decode within 1\%. At batch 1 decode is different: every weight is read once for one multiply-add (over 1.3\,GB of expert weights per
token), so a step is bound by streaming weights and by per-kernel latency, and INT8 arithmetic gives no gain.

\paragraph{Expert layout.} Building the expert tensors by fancy indexing leaves the interleaved gate$|$up block codes and row scales in a strided layout, and a Metal kernel call copies a non-contiguous input whole: about 13\,MB (SC-MXFP4) or 52\,MB (16-wide, four times the block-code bytes) for each of the model's 48 gate$|$up calls. Values are unaffected. In a per-step profile (macOS~27.2, three sessions per row, medians) the expert GEMVs take 12.4\,ms (SC-MXFP4) and 21.1\,ms (16-wide) with the strided layout and 8.8 and 8.2\,ms when contiguous; the full step goes from 29.4 to 25.6\,ms and from 38.3 to 24.0\,ms. Every timing in this paper uses the contiguous layout, which raised Qwen3.5-35B-A3B decode from $0.90\times$ to $0.99\times$. Perplexity on the kernels is the same with either layout (2.511 over 8 windows for 16-wide, 6.052 for SC-MXFP4 on Qwen3.5). Sessions started soon after another process freed 60+\,GB ran up to $1.6\times$ slow in prefill, so Table~\ref{tab:q38speed} uses a 5\,min pause before each session.

\paragraph{GSM8K details.} Lenient answer extraction gives 92.95\% for our W4A8 system on SC-MXFP4, 93.40\% for MLX's W4A16 kernels on the same weights, 92.42\% on MLX's own MXFP4, 95.15\% for the 16-wide format and 95.45\% for the 8-wide format. Paired (strict match): W4A8 vs.\ W4A16 on SC-MXFP4, 17 vs.\ 19 problems solved by only one of the two ($p=0.87$; same answer on 1261 of 1319); SC-MXFP4 vs.\ MLX's MXFP4, both W4A16, 52 vs.\ 28 ($p=0.010$; lenient $p=0.15$); 16-wide vs.\ SC-MXFP4, 52 vs.\ 23 ($p=0.001$), vs.\ MLX's MXFP4, 70 vs.\ 19 ($p<10^{-4}$); 8-wide vs.\ MLX's MXFP4, 73 vs.\ 16 ($p<10^{-9}$; lenient 60 vs.\ 20, $p<10^{-5}$), vs.\ SC-MXFP4 W4A8, 57 vs.\ 22 ($p=1.0\times10^{-4}$), vs.\ 16-wide, 23 vs.\ 17 ($p=0.43$, same answer on 1272 of 1319). All with the model's Metal DeltaNet kernel. Over the first 8 windows, perplexity on our kernels is 2.633 against 2.638 for the reference with BF16 rounding, and 2.511 against 2.521 for the 16-wide format (reference DeltaNet recurrence; with the Metal kernel, 2.647 against 2.640 and 2.511 against 2.514).

\paragraph{Against concurrent INT8 kernels.} oMLX's INT8-activation kernels~\citep{omlx} (release 0.7.1rc1, the
same MLX version) multiply MLX's affine INT4 weights against INT8 activations and correct every 64-element group.
At this model's shapes (random weights, chained calls), our dense projections run at $0.89$--$1.11\times$ their speed at
512 tokens and $1.07$--$1.24\times$ at 2048--8192 tokens; for the routed experts their gate/up kernel, which decodes
weights directly into the matrix unit's register fragments, is faster than our grouped GEMMs, which stage them in
threadgroup memory, although their down projection stays in W4A16 (ours: $0.94$, $0.86$ and $0.76\times$ their speed at
512, 2048 and 8192 tokens). The per-group correction therefore costs well-engineered kernels little; what the
integer-exact formats add is accuracy (with INT8 activations affine INT4 loses $+2.1\%$ on the routed experts,
ours at most $+0.2\%$, Table~\ref{tab:q38}), exactness, and, for SC-MXFP4, compatibility with standard MXFP4 files, and register-level decoding is an
optimization our kernels can adopt. We did not run oMLX end to end: its checkpoint for this model keeps the n-gram
table in memory (30\,GB at 4 bits), which with the weights would take about 120\,GB of the 128\,GB.

\section{Full related work}
\label{app:related}

\paragraph{W4A8 and integer scales.} Foundational 8- and 4-bit LLM quantization methods mitigate activation outliers via mixed-precision decomposition~\citep{llmint8} and preserve salient weight channels or normal distributions for W4A16 execution~\citep{awq,qlora,marlin}. VS-Quant~\citep{vsquant} introduced a coarse float scale times small integer
per-vector scales, the structure of llama.cpp's k-quants and of IQ4\_KS (a float row scale, 7-bit integer block
scales and an INT8 codebook at 4.25 bits)~\citep{llamacpp,iq4ks}; its block scales times its codebook exceed the INT8
range, so its kernels scale each block's partial sum. QServe~\citep{qserve} and DGQ~\citep{dgq} bound
INT4 codes times integer group scales so that the weights are INT8 operands of one GEMM; QQQ~\citep{qqq} and Integer
Scale~\citep{integerscale} apply amplified integer group scales to partial sums. Cross-vendor INT8 matrix units on NVIDIA Ampere/Ada~\citep{nvidia2020a100,nvidia2022ada}, AMD CDNA~\citep{amd2023cdna3}, Intel AMX~\citep{intel2024amx}, Arm SME~\citep{hellosme}, and Apple M5 motivate targeting unmodified INT8 GEMMs. Our format family
(\S\ref{sec:family}) is the INT8-closed instance of this structure with the FP4 codebook instead of INT4, which is
where its accuracy comes from (Table~\ref{tab:formats}).

\paragraph{FP4 weights on integer hardware.} That doubled E2M1 values are integers is used by llama.cpp's MXFP4 and
NVFP4 kernels (an INT8 lookup table, dot products against INT8 activations), by a CPU W4A8 MXFP4 kernel in
ML-SpecQD, by BlockDialect mixed-format execution~\citep{blockdialect}, and by an open vLLM GEMV for AMD GPUs that reads MXFP4 checkpoints unchanged; all apply the E8M0 scale to
each 32-element block's partial sum~\citep{llamacppfp4,mlspecqd,vllmw4a8mxfp4}; in the other direction, FP64 GEMMs
have been emulated on FP4 tensor cores with integer limbs held in E2M1~\citep{hayashi2026ozaki}. ZeroQuant-FP constrains group scales to powers of two
so that casting FP4 to FP8 is a shift~\citep{zeroquantfp}, shared microexponents confine sub-block exponents to a
window~\citep{sharedmicroexp}, and AdaMX reports that over 90\% of output channels have block exponents within a range
of three, which its accelerator exploits with a per-channel bias~\citep{adamx}. We give the exact row-level condition
under which every per-block scale folds into INT8 operands, the quantizer and bound that enforce it, and the
consequence that a standard MXFP4 layer is one INT8 GEMM with one rescale per output.

\paragraph{FP4 formats and accuracy.} MX formats~\citep{ocpmx,rouhani2023microscaling} and NVFP4~\citep{nvfp4} target native FP4
hardware. MR-GPTQ~\citep{mrgptq} combines block rotations with GPTQ to bring MXFP4 close to NVFP4, Four Over
Six~\citep{fourover6} chooses NVFP4 block scales adaptively on existing hardware, and recent scale-search and smoothing methods~\citep{scalesearch,scalesweep,soscale,osfp4} optimize block scales directly against layer-output Hessian error; \citet{chhugani2026mxfp4} add a fractional 128-element partial-sum scale for MXFP4, trading partial-sum work for range. By contrast, M2XFP~\citep{m2xfp}, NxFP~\citep{nxfp}, HiFloat4~\citep{hifloat4}, MixFP4~\citep{mixfp4},
IF4~\citep{if4}, and AdaMX~\citep{adamx} refine block scales or element formats for new hardware rather than integer matrix units; INT v.s.
FP~\citep{intvsfp} and ARCQuant~\citep{arcquant} compare formats under direct casting. Like the round-to-nearest ablations of MR-GPTQ and INT v.s. FP,
we find that block rotations do not help E2M1 formats or NVFP4 without GPTQ, and extend this to W4A8; with GPTQ they help every format we tested, in simulation (Appendix~\ref{app:ablation}).
Rotations~\citep{quarot,spinquant}, smoothing~\citep{smoothquant}, and GPTQ~\citep{gptq} remain compatible with
SC-MXFP4. MoE-specific methods---low-rank expert compensators~\citep{milo}, expert-balanced calibration~\citep{moequant}, mixed-precision routing/expert allocation~\citep{mxmoe}, and compressed-space attention~\citep{rotaryquant}---complement our finding (\S\ref{sec:acc}, \S\ref{sec:q38}) that MoE dense projections hold most of the quantization error and decode bytes, while calibration-free Sinkhorn normalization~\citep{sinq} improves base weight conditioning orthogonally.

\paragraph{LLM inference on Apple Silicon.} MLX~\citep{mlx} and llama.cpp's Metal backend~\citep{llamacpp}
dequantize 4-bit weights to 16 bits inside their kernels. BaseRT~\citep{basertm5} routes prefill GEMMs through the
M5 Neural Accelerators with 4- and 8-bit weights but does not report operand types or activation quantization, and
its engine is distributed as a binary~\citep{basert}. Concurrently with this work, the open-source projects
oMLX~\citep{omlx} and scratchy~\citep{scratchy} added INT8-activation prefill on the M5 for MLX's affine INT4
weights, applying the affine correction at every 64-element group; in the version we compared against (October 2026) oMLX routes the expert gate and up projections of
Qwen3.8-Flash-Next this way and leaves the down projection in W4A16. Since November 2026 oMLX also runs standard MXFP4 experts against INT8 activations, applying each 32-element block's E8M0 scale to that block's partial sum, and INT8 activations at the dense and shared-expert projections~\citep{omlxmx}; Cider~\citep{cider} ran per-channel INT4 weights (the trivial member of our family) against INT8 activations on the M5 earlier in 2026, and open MLX proposals stage weights as INT8 or correct groups inside the kernel~\citep{mlxissue4627,mlxpr4722,mlxpr4802}. Ours differs in folding every block scale into the operands (one exact INT8 GEMM and one rescale per output, no per-block work in the reduction), in the integer-exact formats that this makes possible, which are more accurate than NVFP4, and
in exactness: our integer matrix products are exact, and we measure the accuracy of the
deployed W4A8 path. On Apple's CPU matrix units, INT8 outer products run at $2\times$ the rate of FP16 with FP32
accumulation~\citep{hellosme}.
\citet{benazir2025profiling} identify dequantization as a limiter of quantized inference on Apple Silicon. Systems
that stream weights of models larger than memory from flash~\citep{llminaflash,flashmoe} target decode and are
complementary.

\section{Full limitations}
\label{app:limits}

End-to-end speed is measured on one M5 Max (GPU and CPU matrix units) and one NVIDIA L4 (prefill on Qwen3-4B and Mistral-7B-v0.3; NVIDIA A100 remains measured only on isolated linear layers, Appendix~\ref{app:micro}). Tables mix macOS~26.6 and 27.2 where marked. Activation-aware NVFP4 search without SmoothQuant was measured only on Qwen3-1.7B and 4B (1--1 split), while folded SmoothQuant + \texttt{act-gram} was tested across three dense models (Qwen3-1.7B, Qwen3-4B, and SmolLM3-3B) plus SmoothQuant + RTN on Mistral-7B-v0.3 (Appendix~\ref{app:ablation}). On Qwen3.8-Flash-Next the 16-wide format is 1.1\% behind NVFP4 in perplexity (the 8-wide format is 1.5\% ahead of it),
main results use round-to-nearest without calibration, GPTQ results are simulated on three small models and rotation runs on our kernels only for Qwen3.5-35B-A3B (\S\ref{sec:acc}), and the
125B model is quantized from its FP8 release rather than from BF16. Under llama.cpp's perplexity protocol (Appendix~\ref{app:llamacpp}) its
Q4\_K\_M (about 5 bits per weight) is more accurate than our formats, and SC-MXFP4 also trails its other 4-bit types on the two MoEs. End-to-end gains are bounded by the work outside
the quantized matrices, which is large in hybrid models such as Qwen3.8-Flash-Next ($1.18$--$1.35\times$ end to end
versus $1.33$--$1.65\times$ for its expert layers), and batch-1 decode does not benefit from INT8 arithmetic: it runs at $0.98$--$1.03\times$ MLX's speed with SC-MXFP4 (on the 125B model with MLX's GEMV for the dense projections) and $0.90$--$0.98\times$ with the 8-wide format, and steps of 2 to 32 tokens run slower than MLX's (\S\ref{sec:speed}). The model's Metal kernel for the DeltaNet
recurrence is less precise than its reference implementation, which we use for the perplexity tables (it changes
8-window perplexity by $-0.3$ to $+0.6\%$, Appendix~\ref{app:method}). The
integer-exact formats are no longer standard MXFP4 files. INT8-activation prefill on the M5 has been developed
concurrently in open-source projects for per-channel and affine INT4 weights and for MXFP4 with per-block partial sums~\citep{cider,omlx,scratchy,omlxmx}, and at the kernel level our design
is not faster than theirs (\S\ref{sec:q38}): the advantage of folding the scales is exactness and the formats it
enables, not speed over carefully engineered per-group corrections.

\end{document}